\documentclass[conference]{IEEEtran}

\usepackage[export]{adjustbox}[2011/08/13]

\usepackage[flushleft]{threeparttable} 
\usepackage{booktabs}

\usepackage{graphicx}  
\usepackage[caption=false,font=footnotesize]{subfig}  
 
\usepackage{amsmath,amssymb,amsfonts}   
\usepackage{mathtools}
\usepackage{algorithm}
\usepackage{algorithmic}
\usepackage{textcomp}
\usepackage{xcolor}
\usepackage{array} 
\usepackage{multirow} 
\usepackage{enumerate} 

\usepackage{tabularx}

\makeatletter
\newcommand*\titleheader[1]{\gdef\@titleheader{#1}}
\AtBeginDocument{%
  \let\st@red@title\@title
  \def\@title{%
    \bgroup\normalfont\large\centering\@titleheader\par\egroup
    \vskip1.5em\st@red@title}
}
\makeatother

\title{A Change Point Detection Integrated Remaining Useful Life Estimation Model under Variable Operating Conditions}
\titleheader{This article has been accepted in Control Engineering Practice \\(DOI: https://doi.org/10.1016/j.conengprac.2023.105840).}

\author{
  \IEEEauthorblockN{Anushiya Arunan}
  \IEEEauthorblockA{
    Engineering Product Development\\
    Singapore University of Technology and Design\\
    Email: anushiya\_arunan@mymail.sutd.edu.sg
  }
  \and
  \IEEEauthorblockN{Yan Qin}
  \IEEEauthorblockA{
    Engineering Product Development\\
    Singapore University of Technology and Design\\
    Email: zdqinyan@gmail.com
  }
  
  \and
  \IEEEauthorblockN{Xiaoli Li}
  \IEEEauthorblockA{
    Institute for Infocomm Research\\
    Agency for Science, Technology and Research\\
    Email: xlli@i2r.a-star.edu.sg
  }
  
    \and
  \IEEEauthorblockN{Chau Yuen}
  \IEEEauthorblockA{
    School of Electrical and Electronics Engineering\\
    Nanyang Technological University\\
     Email: chau.yuen@ntu.edu.sg
  }

}

\begin{document}


\twocolumn
\maketitle
\begin{abstract}
By informing the onset of the degradation process, health status evaluation serves as a significant preliminary step for reliable remaining useful life (RUL) estimation of complex equipment. However, existing works rely on \textit{a priori} knowledge to roughly identify the starting time of degradation, termed the change point, which overlooks individual degradation characteristics of devices working in variable operating conditions. Consequently, reliable RUL estimation for devices under variable operating conditions is challenging as different devices exhibit heterogeneous and frequently changing degradation dynamics. This paper proposes a novel temporal dynamics learning-based model for detecting change points of individual devices, even under variable operating conditions, and utilises the learnt change points to improve the RUL estimation accuracy. During offline model development, the multivariate sensor data are decomposed to learn  fused temporal correlation features that are generalisable and representative of normal operation dynamics across multiple operating conditions. Monitoring statistics and control limit thresholds for normal behaviour are dynamically constructed from these learnt temporal features for the unsupervised detection of device-level change points. The detected change points then inform the degradation data labelling for training a long short-term memory (LSTM)-based RUL estimation model. During online monitoring, the temporal correlation dynamics of a query device is monitored for breach of the control limit derived in offline training. If a change point is detected, the device's RUL is estimated with the well-trained offline model for early preventive action. Using C-MAPSS turbofan engines as the case study, the proposed method improved the accuracy by 5.6\% and 7.5\% for two scenarios with six operating conditions, when compared to existing LSTM-based RUL estimation models that do not consider heterogeneous change points. 

\end{abstract}


\section{Introduction}
\label{intro}
Production efficiency and process safety of complex systems are contingent on individual devices operating reliably. Accurate remaining useful life (RUL) estimation is a key enabler of device reliability as condition based, and preventive maintenance can be timely scheduled based on the RUL information. Generally, the RUL of critical assets is defined as the length of time from the current time to end of useful life, i.e., complete failure~\cite{si2011remaining}. An accurate RUL estimation provides crucial guidance for early action to prevent downtime due to unexpected breakdown.

With the advent of Industry 4.0 and the massive amount of data generated from Industrial Internet of Things sensors, research interest in data-driven RUL estimation models has grown rapidly. 
These models are developed to capitalise on the temporal nature of sensor data, and can be broadly grouped into classical statistics-based models and deep learning based approaches. In statistics-based models, Ordóñez \textit{et al.} \cite{ordonez2019hybrid}, for instance, captured time dependence with a combined auto-regressive integrated moving average - support vector machine regression model for RUL estimation of engines. Wang \textit{et al.} \cite{wang2012chmm} proposed a continuous hidden Markov model to extract degradation state information from time sequences to feed their RUL estimation model for milling tools. Zheng \textit{et al.} \cite{zheng2018data} utilised a sliding window approach to input time series sensor data into an extreme learning machine based RUL estimation model. However, some drawbacks of these approaches are the inability and computational impracticality of accounting for long-term time dependencies such as in Markov models \cite{wang2012chmm}, and the lost time dependency information when sliding windows are assumed to be independent of each other \cite{zheng2018data}. 

To address these shortcomings, deep learning approaches have gained popularity in recent years as the backbone for fault diagnosis and RUL estimation. These models have been proposed for a range of critical equipment such as  machine bearings \cite{chen2022mswr, zhu2018estimation, xia2018two, ma2020deep}, cooling systems \cite{wu2021remaining,huang2018remaining}, engines \cite{xiang2021multicellular, zhang2020time,liu2022aircraft, zhang2016multiobjective,mo2023few}, and even batteries \cite{10128148,liu2022towards, qin2021transfer, chen2021remaining}. Particularly, the long-short term memory (LSTM) architecture is of interest compared to standard recurrent neural networks due to its ability to capture both short-term patterns and long-term dependencies in time series via the information-sieving mechanisms of the LSTM’s input, forget, and output gates \cite{hochreiter1997long}.

Despite the superior abilities of LSTM, existing RUL estimation models often use simplifying assumptions or domain knowledge-reliant literature values for modelling the RUL progression through a machine's lifetime \cite{zheng2017long,chen2020machine,huang2019bidirectional, liao2018uncertainty, wu2019weighted}. Generally, critical assets operate normally in the beginning and the onset of degradation only occurs after an uncertain time point, defined as the change point \cite{shi2021dual}. Taking turbofan engines for instance, this non-linear progression of RUL is often represented in a piecewise manner, where the RUL is capped at a constant upper limit during initial operation cycles, and only starts to decrease after some time in operation. The upper RUL limit is typically capped using prior literature values from operational experiments \cite{heimes2008recurrent, zheng2017long,chen2020machine,huang2019bidirectional, liao2018uncertainty}. However, an obvious shortcoming of such an approach is the considerable domain expertise needed for selecting suitable change points. 

There have been a few budding research efforts recognising the need for data-guided change point detection to improve RUL estimation models. For instance, Wu \textit{et al.} \cite{wu2018remaining} utilise a support vector machine-based anomaly detector to identify the change point prior to an LSTM-based RUL estimation. However, their method is evaluated on  only a small test size of 20 engines, and the work focuses solely on late-stage RUL estimation (defined as the last 50 cycles before failure). Meanwhile, Shi and Chehade \cite{shi2021dual} put forward a dual-LSTM model to consider the heterogeneous change points of different devices in their RUL estimation. The first LSTM model classifies if an engine is in a normal or degradation state to detect the change point, while the second LSTM model performs the RUL estimation. Their work similarly focuses on late-stage RUL estimation, and though the RUL estimation performance was promising for cases with single operating conditions, the performance under multiple operating conditions is unknown. Interested readers may also refer to Appendix A for a detailed comparison of these existing works and our current work.

In practice, a device may experience variable and frequently switching operating conditions, resulting in heterogeneous degradation behaviour among different devices. Hence, it is difficult to achieve accurate RUL estimation when individual differences in degradation behaviour are overlooked. Therefore, in this paper, we discuss and address the following two crucial yet unsolved challenges:
\vspace{-0.2em}
\begin{itemize}
  \item Variable operating conditions naturally produce disparate degradation processes, resulting in different change points for individual devices. However, current approaches of degradation modelling still apply a prior knowledge-reliant, fixed representation for all devices of the same type. Particularly, existing works have not investigated how specific change points of individual devices can be identified from each device's degradation behaviour, and utilised for enhancing RUL estimation capabilities.
\vspace{0.2em}
  \item Health status evaluation of whether a device is in normal operation or degradation state is an essential preliminary step for developing reliable RUL estimation models. However, existing degradation modelling studies require well-labelled data to distinguish between different states. As in-depth knowledge of underlying operating conditions and labelled data are not always available, supervised models can be unreliable when extended to devices with different working principles. Thus, there is an urgent need for an unsupervised, generalizable, and data-driven method to account for dynamic degradation behaviours within RUL estimation models.

\end{itemize}

To address these gaps thoroughly and handle the challenges of variable operating conditions, we propose in-depth analysis of local temporal dynamics for health status evaluation and change point detection, prior to an LSTM-based RUL estimation model. Here, the local temporal dynamics is defined as the short-term correlations between a limited number of adjacent past and future lags of sensor measurements. Notably, we successfully extract out generalisable  temporal variations that are representative of normal operation
dynamics across multiple operating conditions, by a novel leveraging of canonical variate analysis (CVA), to automatically detect change points based on significant changes to these temporal variations. Specifically, latent local temporal correlations are learnt and extracted from raw sensor measurements to dynamically construct the monitoring statistics and control limit for the unsupervised detection of device-level change points. The change points then inform the degradation data labelling for training the LSTM-based RUL estimation model, which is now health-status cognizant. Using the trained model, an online query device's change point and RUL can be estimated for early preventive action. Overall, the key contributions of this paper are: 
\begin{enumerate}[i)]
  \item We introduce a novel temporal learning methodology for analysing the latent temporal dynamics of sensor measurements and tracking device degradation progression under variable operating conditions.
  \item We propose a comprehensive and generalisable health status-dependent RUL estimation model, where the RUL estimation of individual devices is enhanced with precise change point detection. Our unsupervised change point detection method circumvents the need for domain expertise and ground-truth based labelling of train data.
  \item We validate the criticality of accounting for heterogeneous change points of individual devices, especially complex devices with multiple operating conditions, by achieving 5.6\% - 7.5\% improvement in RUL estimation accuracy.
\end{enumerate}

The remainder of the paper is organised as follows. Section II describes the preliminaries for a better understanding of the proposed method. Section III outlines the proposed method for change point detection and RUL estimation. Section IV discusses the data, experiments, and results. Finally, Section VI concludes and highlights future research directions.

\section{Preliminary}
\label{prelim}
This section builds the premise for change point integrated RUL estimation and discusses the concept of change point and the standard LSTM model to build the integrated framework.

\subsection{Definition of Change Point}
\begin{figure}[!t]
\centering
\includegraphics[width=0.75\linewidth]{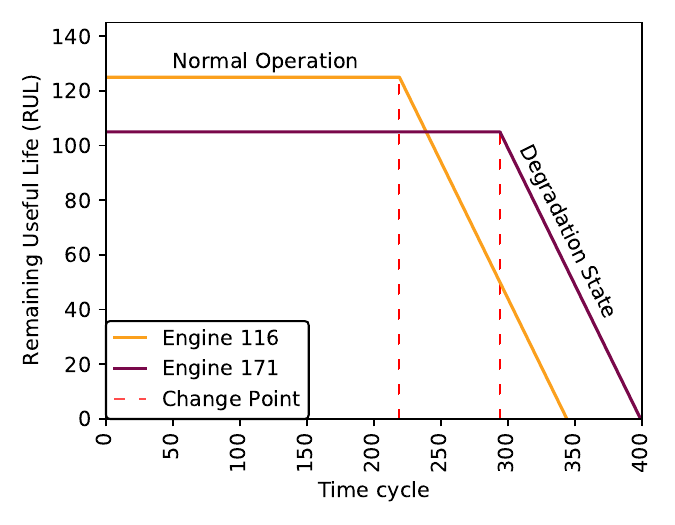}
\vspace{-1.2em}
\caption{Piecewise RUL function showing the relationship between the change point and upper RUL limit value, using randomly selected engines of FD004 in the Commercial Modular Aero-Propulsion System Simulation dataset.}
\label{UpperRUL-piecewise RUL}
\end{figure}

Heimes \cite{heimes2008recurrent} is one of the first influential works to popularise the use of a piecewise-linear function (i.e., constant RUL, followed by a linearly decreasing RUL) to model a device's RUL progression throughout its lifespan. The need for representing a device’s degradation process in a piecewise manner arises because its lifecycle can be broadly divided into two states: a healthy, normally operating state and a degradation state.  During the initial operating cycles before the change point, degradation is often negligible, and it can be reasonably assumed that the RUL remains relatively unchanged (constant) for practical modelling purposes. The device's RUL only starts decreasing distinctively when degradation begins after a yet-to-be determined time point, termed the change point.  As seen in Fig. \ref{UpperRUL-piecewise RUL}, the RUL is capped by an upper limit during normal operation and diminishes only during the degradation state. The change point marks the shift from normal operation to the degradation state. As a side, the RUL is assumed to decrease linearly after the change point in our work, following \cite{heimes2008recurrent, zheng2017long,chen2020machine,huang2019bidirectional, liao2018uncertainty, greenbank2023piecewise}. However, other variants such as a non-linear RUL decay after the change point can also be easily considered depending on the dataset characteristics and domain application.

\subsection{Change Point Integrated RUL Estimation}
The piecewise RUL target label is a key input for RUL estimation models. However, an important difference in the existing approaches of constructing the piecewise labels is that only fixed literature values of the upper RUL limit are available for capping the RUL function, and thus, these values are directly used \cite{zheng2017long,chen2020machine,huang2019bidirectional, liao2018uncertainty, wu2019weighted}. Consequently, the change points of individual devices are not explicitly known or investigated. In contrast, a change point integrated RUL estimation seeks to detect the change point first, and then calculate the unique upper RUL limit for each device following:
\vspace{-0.5 em}
\begin{equation}
{y}_j^{max}=\ k_j^{max}\ -\ k_{j}^{cp}
\end{equation}
\noindent where ${y}_j^{max}$ is the upper RUL limit for device $j$, given its change point, $k_j^{cp}$ and its maximum lifespan $k_j^{max}$.

The piecewise degradation data constructed from the learnt change points is fed to the LSTM model discussed next to form the change point integrated RUL estimation framework.

\subsection{LSTM Model for RUL Estimation}
An LSTM-based RUL estimation models the non-linear relationship between input sensor data and the piecewise RUL labels (i.e., degradation data). A basic LSTM unit is a cell with three gates (input, forget and output) to sieve information flow through the cell. LSTM is a recurrent network as both its cell state $\mathbf{c}_k$ and hidden state, $\mathbf{h}_k$ at time $k$ holds the memory from the previous cell state, $\mathbf{c}_{k-1}$, hidden state $\mathbf{h}_{k-1}$ and input $\mathbf{x}_k$ \cite{hochreiter1997long}. The predicted RUL $\widehat{\mathbf{y}}_k$ is determined by $\mathbf{h}_k$. For interested readers, further details of the standard LSTM architecture can be found in \cite{hochreiter1997long}. 

\section{Change Point Detection Integrated Model for Remaining Useful Life Estimation}
\label{change point method}
\begin{figure*}[!t]
\includegraphics[width=\textwidth,center]{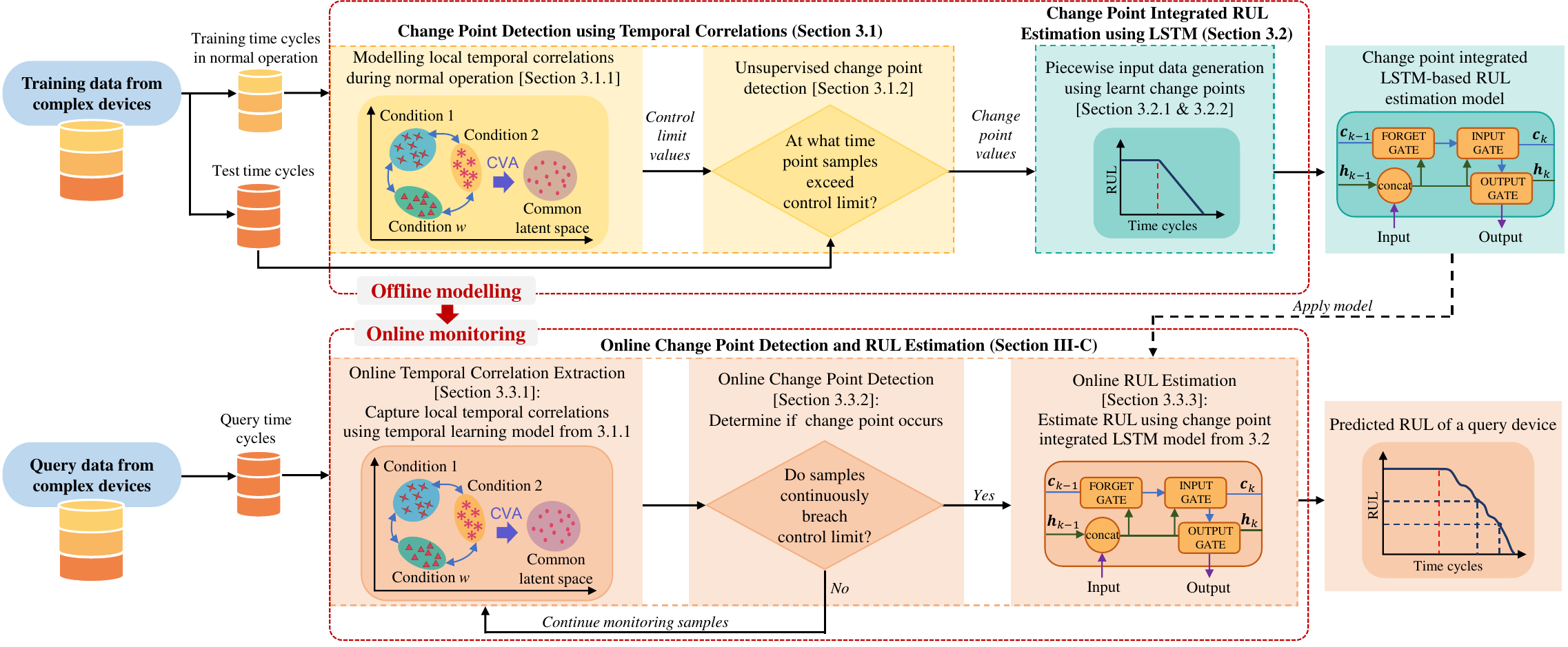}
\vspace{-1.7em}
\caption{Overall structure of proposed change point detection-based RUL estimation method.}
\label{processflow}
\end{figure*}

The change point detection integrated RUL estimation model can be divided into offline modelling and online monitoring. In offline modelling, a change point detection model first analyses the local temporal dynamics of sensor measurements to learn the start time of degradation, i.e., the change point. Then, the learnt change points are utilised to calculate upper RUL limit values for transforming the RUL labels of a device as a piecewise function. With the transformed labels, an LSTM model is developed and trained. During online monitoring, a query device is monitored for the occurrence of a change point and its RUL is estimated with the offline trained models. The steps of the proposed methodology, from change point detection to RUL estimation, are detailed below, and summarised in Fig. \ref{processflow}.

\subsection{Change Point Detection using Temporal Correlations}
Conventionally, CVA is used for finding associations between two multivariate datasets, where the multiple variables in each dataset are considered as a whole. For multiple variables to be considered as a whole, the variables are transformed via appropriate linear combinations (i.e., projections) to a fused latent variable, termed the canonical variate \cite{odiowei2009nonlinear}. In this paper, we capitalise CVA's association finding ability in an innovative way to create an in-depth temporal dynamics monitoring model for detecting degradation change points.

\subsubsection{Modelling of Local Temporal Correlations during Normal Operation}
The procedure for building and training the CVA-based temporal learning model are outlined next. A training set of sensor data during normal operation is required to calculate the canonical variate matrices, health monitoring statistics, and a Control Limit (CL), which acts as an upper threshold for the monitoring statistics. The monitoring statistics of any new test data can be compared against the CL value previously calculated with the normal operation training data to detect statistically significant breaches above the CL, and consequently, the change point at which degradation begins.

To prepare the training data of each device for  capturing the significant latent temporal correlations amongst the sensor measurements, we start with the feature matrix of sensor data obtained during normal operation, $\mathbf{X}\in\mathbb{R}^{m \times N}$, consisting of $m$ sensor variables (e.g., temperature, pressure) and $N$ time series observations. To relate the temporal correlations of multiple sensor data, the past and future vectors, $\mathbf{x}_{p, k}$ and  $\mathbf{x}_{f, k}$ are formed by expanding each sensor measurement at time cycle, $k$ by $p$ past lags and $f$ future lags:
\vspace{-0.27em}
\begin{equation}
\begin{aligned}
\mathbf{x}_{p, k}&=\left[\begin{array}{lll}
\mathbf{x}_{k-1},\mathbf{x}_{k-2}, \ldots, \mathbf{x}_{k-p}
\end{array}\right]^{\mathrm{T}} \in \mathbb{R}^{m p}\\
\mathbf{x}_{f, k}&=\left[\begin{array}{lll}
\mathbf{x}_{k},\mathbf{x}_{k+1}, \ldots, \mathbf{x}_{k+f-1}
\end{array}\right]^{\mathrm{T}} \in \mathbb{R}^{m f}
\end{aligned}
\end{equation}
where $p$ and $f$ represent the number of lags of sensor measurements used for the temporal correlation modelling.

The final step of training data construction is concatenating $\mathbf{x}_{p, k}$ and $\mathbf{x}_{f, k}$ vectors in the variable-wise direction to form the comprehensive past and future matrices, $\mathbf{X}_{p}$ and $\mathbf{X}_{f}$:
\vspace{-0.27em}
\begin{equation}
\begin{aligned}
\mathbf{X}_{p}&=\left[\mathbf{x}_{p, p+1}, \mathbf{x}_{p, p+2}, \ldots, \mathbf{x}_{p, p+\tilde{N}}\right] \in \mathbb{R}^{m p \times \tilde{N}}\\
\mathbf{X}_{f}&=\left[\mathbf{x}_{f, p+1}, \mathbf{x}_{f, p+2}, \ldots, \mathbf{x}_{f, p+\tilde{N}}\right] \in \mathbb{R}^{m f \times \tilde{N}}
\end{aligned}
\end{equation}
where $p$ = $f$ and $\tilde{N}$ = $N$ – $f$ – $p$ +1 for $N$ observations.

The data in  $\mathbf{X}_{p}$ and $\mathbf{X}_{f}$ are standardised to zero mean and unit variance with respect to each sensor variable to prevent large values from skewing subsequent calculations. The transformation matrices required to calculate the canonical variates are derived from the singular value decomposition (SVD):
\begin{equation}
\mathbf{\Sigma}_{f f}^{-1 / 2} \mathbf{\Sigma}_{f p} \mathbf{\Sigma}_{p p}^{-1 / 2}=\mathbf{U D V}^{\mathrm{T}} \in \mathbb{R}^{m f \times m p}
\end{equation}
where $\mathbf{\Sigma}_{f f}^{-1 / 2}$ and $\mathbf{\Sigma}_{p p}^{-1 / 2}$ are the covariance matrices of $\mathbf{X}_{f}$ and $\mathbf{X}_{p}$ respectively, and  $\mathbf{\Sigma}_{f p}$ is the cross-covariance matrix of  $\mathbf{X}_{f}$ and $\mathbf{X}_{p}$; $\mathbf{U}$ and $\mathbf{V}$ are the left and right singular matrices respectively, and $\mathbf{D}$ is a diagonal matrix of non-negative singular values.

The system canonical variates, i.e., the dominant variates with large correlations, $\mathbf{Z}$ and the residual variates with low correlations, $\mathbf{E}$ are calculated as the linear combinations of $\mathbf{X}_{p}$ by applying the respective transformation matrices, $\mathbf{V}_{r}^{\mathrm{T}} \mathbf{\Sigma}_{p p}^{-1 / 2}$ and $\left(\mathbf{I}-\mathbf{V}_{r} \mathbf{V}_{r}^{\mathrm{T}}\right) \mathbf{\Sigma}_{p p}^{-1 / 2}$ to $\mathbf{X}_{p}$:
\begin{equation}
\mathbf{Z}=\mathbf{V}_{r}^{\mathrm{T}} \mathbf{\Sigma}_{p p}^{-1 / 2} \mathbf{X}_{p} \in \mathbb{R}^{r \times \tilde{N}}
\end{equation}
\begin{equation}
\mathbf{E}=\left(\mathbf{I}-\mathbf{V}_{r} \mathbf{V}_{r}^{\mathrm{T}}\right) \mathbf{\Sigma}_{p p}^{-1 / 2} \mathbf{X}_{p} \in \mathbb{R}^{m p \times \tilde{N}}
\end{equation}
where $r$ is the number of system canonical variates.

The changes in canonical variates at each time cycle $k$ is quantitatively measured by two health monitoring statistics, which are the Hotelling $T^2$ statistic and Squared Prediction Error $Q$  statistic. The $T^2$ and $Q$ statistics are complementary as $T^2$ captures the total variation of the system canonical variates, while the $Q$ statistic measures the variations of errors in the residual space:
\vspace{-0.35em}
\begin{equation}
T_{k}^{2}=\sum_{i=1}^{r} z_{i, k}^{2}
\vspace{-0.3em}
\end{equation}
\vspace{-0.2em}
\begin{equation}
Q_{k}=\sum_{i=1}^{m p} \varepsilon_{i, k}^{2}
\vspace{-0.35em}
\end{equation}
where $z_{i, k}$ and $\varepsilon_{i,k}$ are the elements in row $i$ and column $k$ of the respective canonical variate matrices $\mathbf{Z}$ and $\mathbf{E}$.

Lastly, the statistically significant CL values for the health monitoring statistics are calculated. For instance, at a statistical significance level of $\alpha$ = 0.99, the CL establishes an upper threshold value, below which 99\% of $T^2$ and $Q$ statistic sample values will fall during normal operation. To calculate the CL of $T^2$ and $Q$, their probability density functions has to be established. As non-linear systems may not follow a Gaussian distribution, the probability distribution of $T^2$ and $Q$ is modelled using Kernel Density Estimation following \cite{odiowei2009nonlinear}. The CL values for $T^2$ and $Q$, defined as ${CL}_{T^2}$ and ${CL}_Q$, are then obtained by solving $P\left(T^2<\ {CL}_{T^2}\right)=\ \alpha$ and $P\left(T^2<\ {CL}_Q\right)=\ \alpha$, respectively.

\subsubsection{Unsupervised Change Point Detection}
With the CL values calculated using training data from normal operation, we can assess the remaining sensor data $\mathbf{X}^{test} \in \mathbb{R}^{m \times N}$ of a device, containing time series sensor data from normal operation and degradation states, to identify its change point. The change point detection strategy is summarised in Algorithm 1, and the detailed steps are discussed henceforth.

\begin{algorithm}[t]
\caption{Unsupervised change point detection}
\footnotesize
\algsetup{linenosize=\footnotesize}
\begin{algorithmic}[1]
\renewcommand{\algorithmicrequire}{\textbf{Input:}}
\renewcommand{\algorithmicensure}{\textbf{Output:}}
\newcommand{\algorithmicbreak}{\textbf{break}}
\newcommand{\BREAK}{\STATE \algorithmicbreak}
\REQUIRE Test data, $\mathbf{X}^{test}$ from time, $\tau$ till end of life, $k^{max}$ \\ 
\hspace{3.6mm} Control limits from normal operation, $CL_{T^{2}}$ and $CL_{Q}$ 
\ENSURE Change points, $k_{T^{2}}^{c p}$ and $k_{Q}^{c p}$
\\ \textit{Initialise sampling time, $k$ = $\tau$}
 \STATE Construct $\mathbf{X}_{p}^{test}$ using Eq. (3)
 \STATE Determine $\mathbf{Z}^{test}$ and $\mathbf{E}^{test}$ using Eqs. (9) and (10) 
 \FOR {$k = \tau$ to $k^{max}$}
 \STATE Calculate $T_{k_{-} test}^{2}$ and $Q_{k_{-} test}$ using Eqs. (11) and (12)
 \ENDFOR
\STATE Determine change point, $k_{T^{2}}^{c p}$ based on $T_{k_{-} test}^{2}$:
\STATE Initialise $k = \tau$
\WHILE {($k \leq k^{max}$)}
 \IF {$T_{k_{-} test}^{2}(k) \geq C L_{T^{2}} \text { for all } k \text{ till } k^{max}$}
 \RETURN $k = k_{T^{2}}^{c p}$
 \BREAK
 \ELSE
 \STATE $k = k+1$
 \ENDIF
\ENDWHILE
\STATE Repeat steps (7) through (15) to determine change point, $k_{Q}^{c p}$ based on $Q_{k_{-} test}$ using corresponding control limit $CL_{Q}$
\end{algorithmic} 
\end{algorithm}

First, the lagged past matrix $\mathbf{X}_{p}^{test}$ is constructed following Eq. (3). Next, similar to the approach for training data in Eqs. (5) through (8), the canonical variates and monitoring statistics $T^2$ and $Q$ are calculated for the test data as follows:
\begin{equation}
\vspace{-0.3em}
\mathbf{Z}^{test}=\mathbf{V}_{r}^{\mathrm{T}} \mathbf{\Sigma}_{p p}^{-1 / 2} \mathbf{X}_{p}^{test} \in \mathbb{R}^{r \times \tilde{N}}
\vspace{-0.3em}
\end{equation}
\vspace{-0.4em}
\begin{equation}
\mathbf{E}^{test}=\left(\mathbf{I}-\mathbf{V}_{r} \mathbf{V}_{r}^{\mathrm{T}}\right) \mathbf{\Sigma}_{p p}^{-1 / 2} \mathbf{X}_{p}^{test} \in \mathbb{R}^{m p \times \tilde{N}}
\vspace{-0.6em}
\end{equation}
\vspace{-0.3em}
\begin{equation}
T_{k_{-} test}^{2}=\sum_{i=1}^{r} z_{i, k_{-} test}^{2}
\end{equation}
\vspace{-0.25em}
\begin{equation}
Q_{k_{-} test}=\sum_{i=1}^{m p} \varepsilon_{i, k_{-} test}^{2}
\end{equation}
where $z_{i,k_{-}test}$ and $\varepsilon_{i,k_{-}test}$ are elements in row $i$ and column $k_{-}test$ of the respective matrices $\mathbf{Z}^{test}$ and $\mathbf{E}^{test}$.

The monitoring statistics $T^2$ and $Q$ of the test data are compared against the control limits ${CL}_{T^2}$ and ${CL}_Q$ calculated previously during normal operation. Monitoring statistics values below the control limit indicate normal operations, while a continuous breach above the CL threshold indicates some fault development and a shift into a degradation state. In practice, there is typically a time  period, where the monitoring statistics fluctuate above and below the CL threshold before the engine actually starts to degrade. We name this period the “transition period”. Thus, to ensure that an accurate onset of degradation that is past the transition period is captured, the change point, $k^{cp}$ is defined as the first time point at which a continuous and consistent breach above the CL starts. The change points are deemed to occur when either the $T^2$ or $Q$ monitoring statistics continuously exceed their respective CL values, and they are determined by solving for $k_{T^2}^{cp}$ and $k_Q^{cp}$:
\vspace{-0.2 em}
\begin{equation}
T_{k_{-} test}^{2}\left(k_{T^{2}}^{c p}\right) \geq C L_{T^{2}} \quad \forall \quad k \in\left[k_{T^{2}}^{c p}, k^{m a x}\right]
\end{equation}
\vspace{-1.3 em}
\begin{equation}
Q_{k_{-} test}\left(k_{Q}^{c p}\right) \geq C L_{Q}        
\quad \forall \quad k \in\left[k_{Q}^{c p}, k^{m a x}\right]
\end{equation}
where $k_{T^2}^{cp}$ and $k_Q^{cp}$ are the change points based on $T^2$ and $Q$ statistics respectively, and $k^{max}$ is the device lifespan.

\subsection{Change Point Integrated RUL Estimation using LSTM}
This section discusses how the learnt change points from Section III-A2 are holistically utilised to enhance the quality of the input data used for training our change point integrated RUL estimation model. The input data consists of two interconnected components, the sensor features and the ground truth RUL labels, which have to be considered concurrently when accounting for the change points within the modelling process. We detail this change point-informed generation of the RUL labels and pre-processing of sensor features in the two subsequent sections.

\subsubsection{Change Point-informed RUL Label Generation}
To generate the RUL labels, we take inspiration from Heimes’s seminal experiments \cite{heimes2008recurrent}  to adopt a piecewise-linear model (i.e., constant RUL prior to the start of degradation, followed by a linearly decreasing RUL till complete failure). In most existing papers on turbofan engines for instance, a constant value of 130 \cite{zheng2017long,xiang2021multicellular} is recommended as the upper RUL limit for all devices, even under variable operating conditions. In contrast, learning the change point of individual devices allows for the unique RUL upper limit to be determined for each device according to Eq. (1). Fig. \ref{UpperRUL-piecewise RUL} illustrates the relationship between the change point and the upper RUL limit. For example, Engine 116 of the FD004 dataset in the Commercial Modular Aero-Propulsion System Simulation (C-MAPSS) turbofan engine degradation dataset \cite{saxena2008damage} has a maximum lifespan of 344 cycles and a change point at around 240 cycles. Therefore, its upper RUL limit will be 104.

After the change point, the choice of a linear or non-linear model for the RUL labelling depends on the equipment type and the characteristics of the generated data. In our work, we assume a linear decay of the RUL after the change point based on the well-studied suitability \cite{ zheng2017long,chen2020machine,huang2019bidirectional, liao2018uncertainty} of this assumption for our downstream case study. However, it should be noted that our data-driven change point detection methodology is designed to detect departures from normal operation behaviour solely from the sensor data's temporal dynamics. It does not assume any underlying model for the RUL decrease after the change point, and thus, it can be easily applied to other domains and datasets experiencing a non-linear RUL decay after the change point.

\subsubsection{Change Point-informed Sensor Data Standardisation}
Feature scaling is a standard data pre-processing procedure required before inputting features to an LSTM model. This prevents variables with relatively large values from dominating and skewing model training and convergence. A popular standardisation technique is Z-score normalisation \cite{zheng2017long}, where each sensor variable is scaled to have zero mean and unit variance.

In our work, we enhance the Z-score normalisation process with the learnt change points to perform a piecewise standardisation of the sensor data. First, the mean and standard deviation the of sensor data before the change point (i.e., data in normal operation) are calculated, and these values are used to standardise the entire sensor data of both train and test devices. 
This normal operation based standardisation allows the sensor data variations experienced during the degradation state to be better contrasted and amplified against normal operation variations. 

\begin{figure}[!t]
\centering
\includegraphics[width=0.85\linewidth]{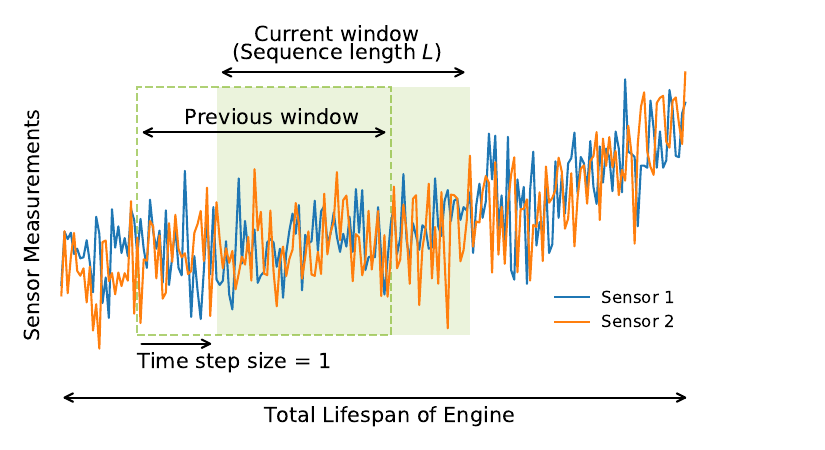}
\caption{Data segmentation using sliding window method with two randomly selected sensor signals shown as examples.}
\label{slidingwindow}
\end{figure}

The aforementioned change point integrated feature data is then segmented into smaller sequences of length $L$ using a sliding window approach, as shown in Fig. \ref{slidingwindow}. The window is shifted through the entire feature data by a step size of 1 at a time. This generates smaller sequence segments, which are consecutively fed into an LTSM network, together with the piecewise-constructed RUL labels, to model the relationship between the feature data and the RUL. The proposed LSTM model consists of 3 stacked layers to increase its capacity to learn important, latent dependencies in the change point-informed input data that can aid accurate RUL estimation. The model is trained to minimise the loss function, the mean squared error of the predicted RUL and true RUL labels, and subsequently, the trained model can be utilised to predict the RUL of any new query device.

\subsection{Online Change Point Detection and RUL Estimation}
With the well-trained change point detection and RUL estimation models, any query sample $\mathbf{X}^{query}$ can be monitored for change point detection and RUL estimation. The online monitoring scheme is summarised stepwise below.

\subsubsection{Online Temporal Correlation Extraction}
Given a query sample $\mathbf{X}^{query} \in \mathbb{R}^{m \times N}$ from a device, its past lagged matrix is constructed using Eq. (3). Next, the local temporal correlation features are extracted for the calculation of $T_{k}^{2}$ and $Q_{k}$ monitoring statistics following Eqs. (9) to (12). 

\subsubsection{Online Change Point Detection}
The calculated $T_{k}^{2}$ and $Q_{k}$ statistics are checked against their respective control limits, ${CL}_{T^2}$ and ${CL}_Q$, established in Section III-A1, to assess if the queried sample falls under normal operation. If the $T_{k}^{2}$ and $Q_{k}$ statistics remain below the CL, the device is operating normally. It is continued to be monitored with no further action. In contrast, if the $T_{k}^{2}$ or $Q_{k}$ statistics continuously exceed the CL, the device is no longer  operating normally. To safely conclude this shift from normal operation to degradation state, the number of time cycles $\lambda$ that the $T_{k}^{2}$ or $Q_{k}$ statistics need to continuously breach the CL should be at least as high as the maximum number of consecutive time cycles the breach occurs for during normal operation and, in any transition period to the degradation state:
\vspace{-0.2 em}
\begin{equation}
{\lambda}= \underset{\Delta {k}}{\arg\max}\left\{\begin{array}{cc}
T_{k}^{2} \geq C L_{T^{2}} \hspace{2mm} \forall \hspace{2mm} k \in\left[k, k + \Delta {k}\right]\\
Q_{k} \geq C L_{Q}         \hspace{2.7mm} \forall \hspace{2mm} k \in\left[k,  k+\Delta {k}\right]
\end{array}\right.
\end{equation}
\noindent where $\Delta {k}$ is the largest increment in time period that the CL breach is sustained for. Following the breach, the change points $k_{T^2}^{cp}$ and $k_Q^{cp}$are detected according to Eqs. (13) and (14). 

On a related note, there could be other domain applications in practice, where detection of multiple change points is  desired. In such cases, the value of $\Delta {k}$ can be appropriately tweaked based on observed normal operation behaviour to fine-tune the sensitivity of the system to detect multiple change points.

\subsubsection{Online RUL Estimation}
If a change point is detected, this indicates the onset of degradation, and an RUL estimation is necessary for planning preventive maintenance. The feature data of the queried sample is processed by piecewise standardisation and sliding window based sequence segmentation before it is fed into the LSTM-based RUL estimation model.

At this juncture, it should also be highlighted that, in industrial processes, data drifts (i.e., changes in data distribution of incoming query data) may gradually occur over time due to factors such as replacement of ageing equipment or changes in operational procedures. As the change point detection model is designed to generalise well over variable operating conditions, we expect our change point integrated RUL estimation model to be reasonably robust against minor data drifts in the near-term.

However, over the long term, substantial data drifts could occur, thus, it is prudent to establish a data and model monitoring pipeline. Incoming test data should be periodically assessed for possible changes in data distributions through standard statistical approaches (e.g., Kolmogorov-Smirnov test \cite{massey1951kolmogorov}, Kullback Leibler divergence \cite{kullback1951information}, etc.). If substantial data drifts occur, the CLs may no longer be representative of the new normal operating conditions. This can be easily remedied offline by retraining or updating the original model with the newly available data to learn the latest normal operating conditions or new fault patterns.

\section{Experiments and Results}
\label{exp and results}
This section assesses the change point detection and RUL estimation performance of the proposed temporal learning model, using the benchmark C-MAPSS turbofan engine degradation dataset \cite{saxena2008damage}. For a fair comparison, performance evaluation is carried out by comparing against LSTM-based deep learning models \cite{zheng2017long,chen2020machine,huang2019bidirectional, liao2018uncertainty, wu2019weighted}.

\begin{table*}[!t]
\newcolumntype{P}[1]{>{\centering\arraybackslash}p{#1}}
\renewcommand{\arraystretch}{1.3}
\tiny
\caption{Descriptive Summary of C-MAPSS Dataset.}
\vspace{-1.2em}
\label{Table_data_summary}
\centering
\begin{center} 
\begin{threeparttable}
\begin{tabular}{m{1.5cm}P{2.7cm}P{2.7cm}P{3cm}P{3cm}}
\hline
\hline
Dataset & \textbf{FD001} & \textbf{FD002} & \textbf{FD003} & \textbf{FD004}\\
\hline
Operating conditions & 1 & 6 & 1 & 6\\
\hline
No. of train engines & 100 & 260 & 100 & 249\\
\hline
No. of test engines & 100 & 259 & 100 & 248\\
\hline
\vspace{0.5mm}Fault components & High pressure compressor & High pressure compressor & High pressure compressor and fan & High pressure compressor and fan \\
\hline
\multirow{6}{*}{\vspace{1.85mm}Distribution of} & \multirow{6}{*}{$\raisebox{-0.95\height}{\includegraphics[scale=0.175]{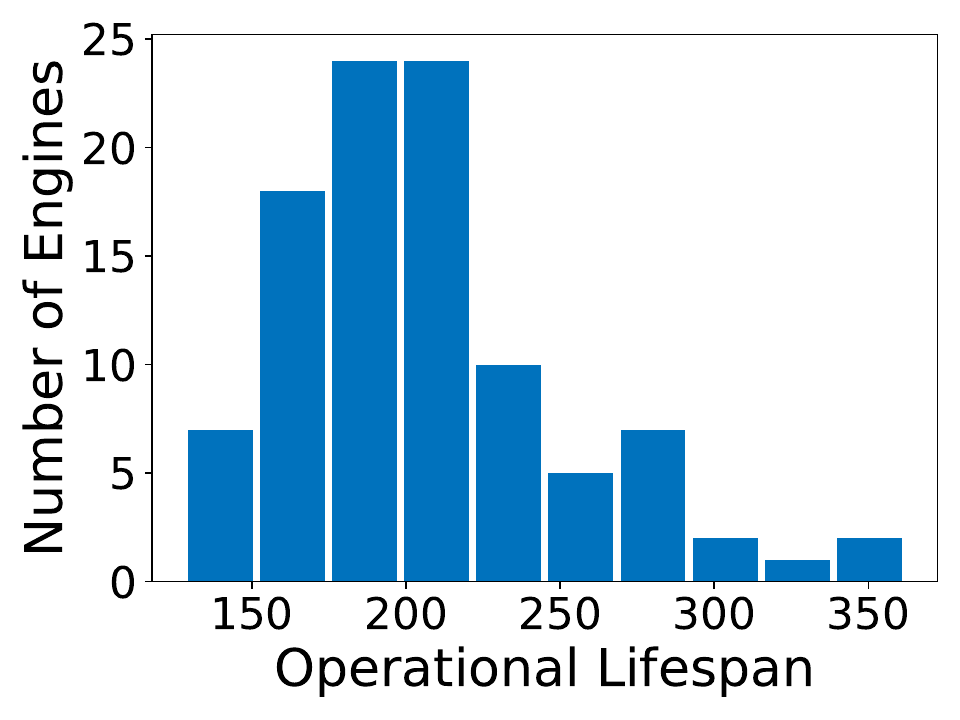}}$} & \multirow{6}{*}{$\raisebox{-0.95\height}{\includegraphics[scale=0.175]{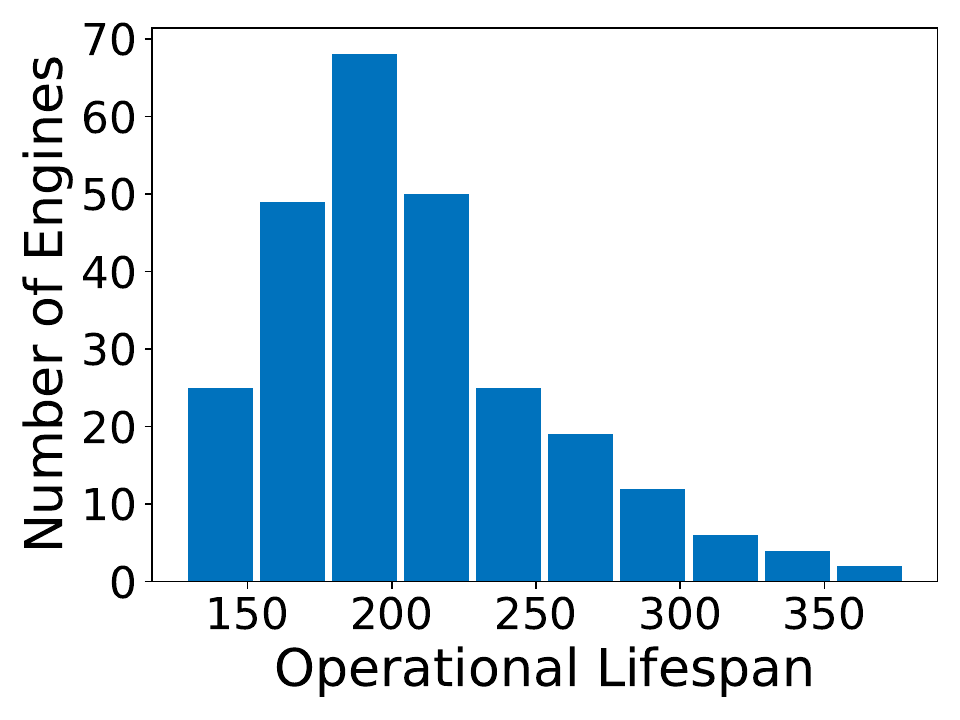}}$} & \multirow{6}{*}{$\raisebox{-0.95\height}{\includegraphics[scale=0.175]{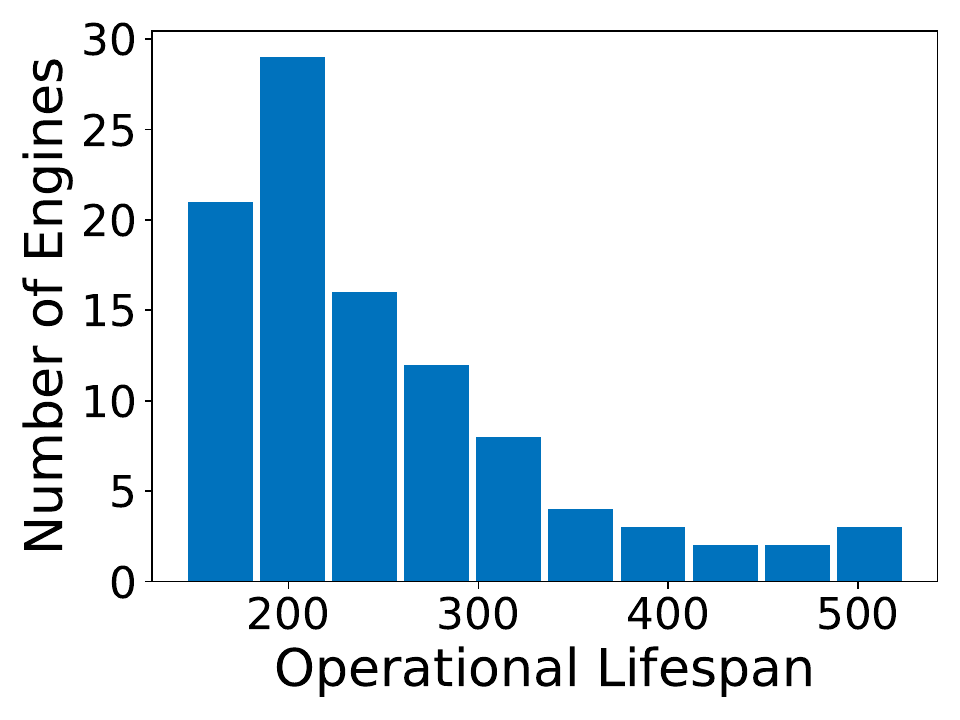}}$} & \multirow{6}{*}{$\raisebox{-0.95\height}{\includegraphics[scale=0.175]{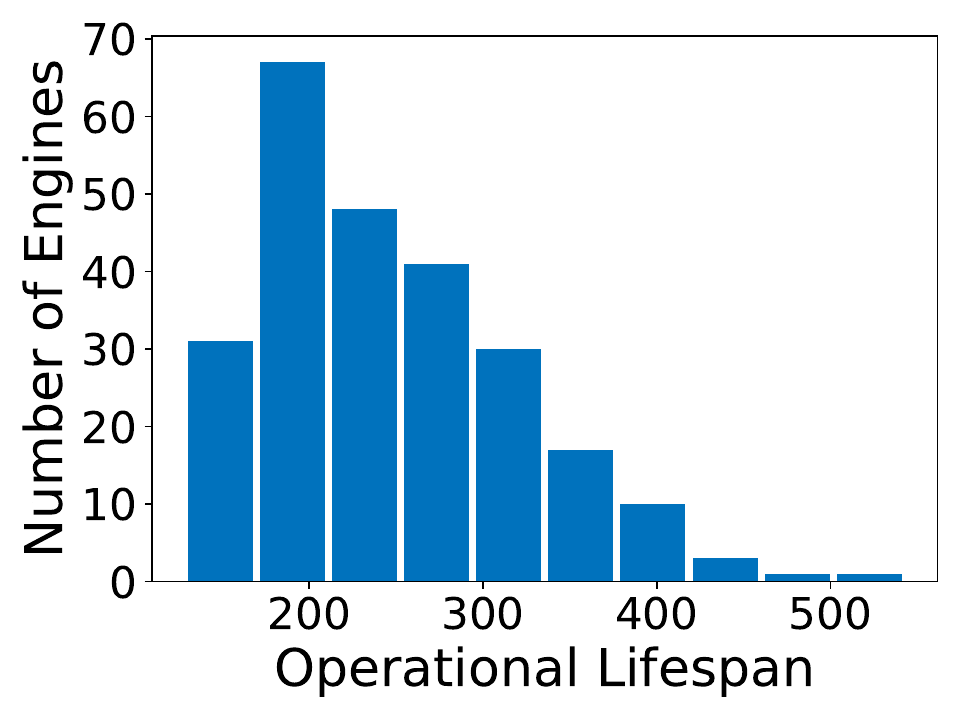}}$} \\
 & & & & \\
 & & & & \\
\vspace{0.01mm}train engine lifespan & & & & \\
 & & & & \\
 & & & & \\
& & & & \\
 & & & & \\
\hline
\hline
\end{tabular}
\end{threeparttable}
\end{center}
\end{table*}

The benchmark C-MAPSS turbofan engine degradation dataset consists of 4 sub-datasets, FD001 to FD004. Each sub-dataset has a varying number of engines, fault modes, and operating conditions. Within each sub-dataset, there is a further division into train and test engines. The dataset is summarised in Table \ref{Table_data_summary}. FD001 is the simplest sub-dataset with engines experiencing 1 operating condition and 1 fault mode, while FD004 is the most complex with 6 operating conditions and 2 fault modes. In the dataset, there are 21 sensor variables (e.g., temperature, pressure) recorded for each operational time cycle of the engine. For the train engines, the time series sensor data are collected from normal operation until system failure. For the test engines, the data are available up to only some random time before failure. The test engines are used to predict the RUL, i.e., number of remaining operational cycles before failure. As the train engine dataset is a time series of operational cycles from normal operation until failure, the maximum number of operational cycles (lifespan) of each engine can be deduced. For instance, in FD001, the lifespan of engines ranged from 128 to 362 cycles.

\begin{figure}[!t]
\centering
\includegraphics[width=\linewidth]{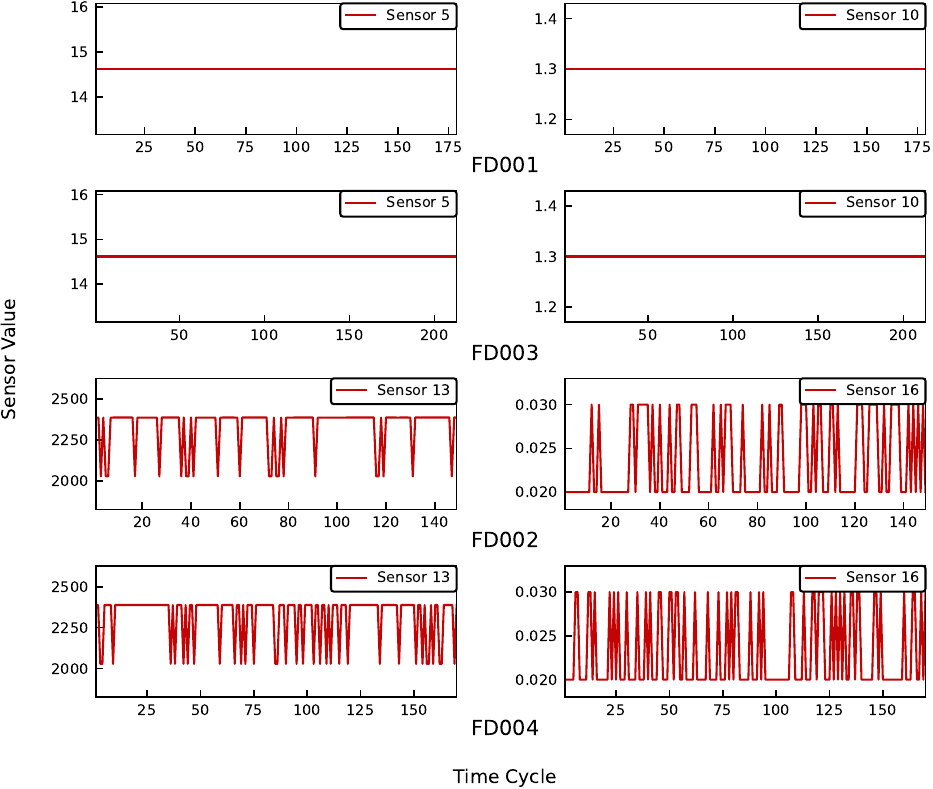}
\vspace{-2em}
\caption{Sample of uninformative sensor readings with constant values for randomly selected engines from FD001 and FD003, and with erratic, range-bound patterns for randomly selected engines from FD002 and FD004.}
\label{sensorplot}
\end{figure}

\subsection{Change Point Detection using Temporal Correlations}
Existing benchmark studies on turbofan engine degradation have extensively discussed which sensors of the C-MAPSS dataset should be selected as features for model training. According to these literature \cite{chen2020machine}, \cite{wu2018remaining, shi2021dual}, only sensor signals with either increasing or decreasing trends should be selected for model development. Sensors that exhibit erratic trends or remain constant over time do not provide useful information about the degradation process and, therefore, can be excluded as features. To guide our sensor selection process, we employ knowledge from existing literature and our own testing with exploratory, trend-checking  plots of the sensor readings. Among the four datasets, FD001 and FD003’s sensor signal patterns are similar because these datasets only contain engines experiencing a single operating condition. For FD001 and FD003, sensors 1, 5, 6, 10, 16, 18, and 19 are excluded as features because these sensor readings largely  remain constant with zero variation, as shown in the representative sensor signal plots of Fig. \ref{sensorplot}. On the other hand, datasets FD002 and FD004 consist of engines experiencing multiple operating conditions. Hence, their sensor signal patterns, as reasonably expected, differ from that of FD001 and FD003. For FD002 and FD004, sensors 10, 13, 16, 18, and 19 are dropped from being features because they have erratic, range-bound measurements with no obvious increasing or decreasing trends, as shown in the representative sensor signal plots of Fig. \ref{sensorplot}.

\subsubsection{Modelling of Local Temporal Correlations during Normal Operation}
To recap, the change point detection model has to be first trained on normal operation data to be able to detect statistically significant deviations from normal operation and degradation change points for new test samples that have both normal operation and degradation conditions. 

For model training, the train engine dataset is used. As this dataset, naturally, does not label sensor data by whether it is from normal operation or not, some reasonable assumptions are needed. First, engines with relatively lengthy lifespan of at least 200 operational cycles are selected to ensure there is sufficient time series data for the training, validation, and testing phases. Interested readers may refer to Appendix B for details leading to the choice of 200 cycles as the minimum lifespan of train engines. This subset of engines still represents a sizeable dataset for the development of a robust change point detection model as nearly 50\% or more of the train engines have a lifespan of at least 200 cycles (refer to the distribution of train engine lifespan in Table \ref{Table_data_summary}). For the remaining engines with a lifespan of less than 200 cycles, whose change points are not determined by the detection model due to data size insufficiency, we adopt the commonly used literature value of 130 cycles \cite{zheng2017long} as the RUL upper limit for the piecewise modelling of RUL target labels in the later Section IV-B1.

For the engines with a lifespan of at least 200 cycles, sensor data from first 60 operational cycles of each engine are assumed to be from normal operation. The lagged sensor data, using $p$ past lags and $f$ future lags, are computed using Eqs. (2) and (3). Typically, $p$ and $f$ have the same values and their optimal values, i.e., statistically significant number of time lags are determined by comparing the autocorrelation function of the summed squares of the measurements in the training data against a certain confidence interval\cite{ruiz2015statistical}. However, due to the limited size of normal operation data available in the C-MAPSS dataset, a smaller value of 2 is used for the number of $p$ and $f$ lags in our construction of past and future matrices. Hence, the temporal correlations extracted are termed local. The lagged sensor data matrices are transformed via Eqs. (5) through (8) to calculate the monitoring statistics, $T^2$ and $Q$. The optimal number of system canonical variates $r$ to calculate $T^2$ and $Q$ is determined based on the downstream RUL estimation performance, and it is discussed in greater detail in Section IV-B4. For FD001 to FD003, $r=15$ resulted in the best RUL estimation performance, whereas, for FD004, which has multiple fault modes and operating conditions, $r=21$ yielded the best RUL estimation.

\begin{figure}[!t]
\vspace{-1.2 em}
\centering
\subfloat[Train cycles]{\includegraphics[width=0.5\linewidth]{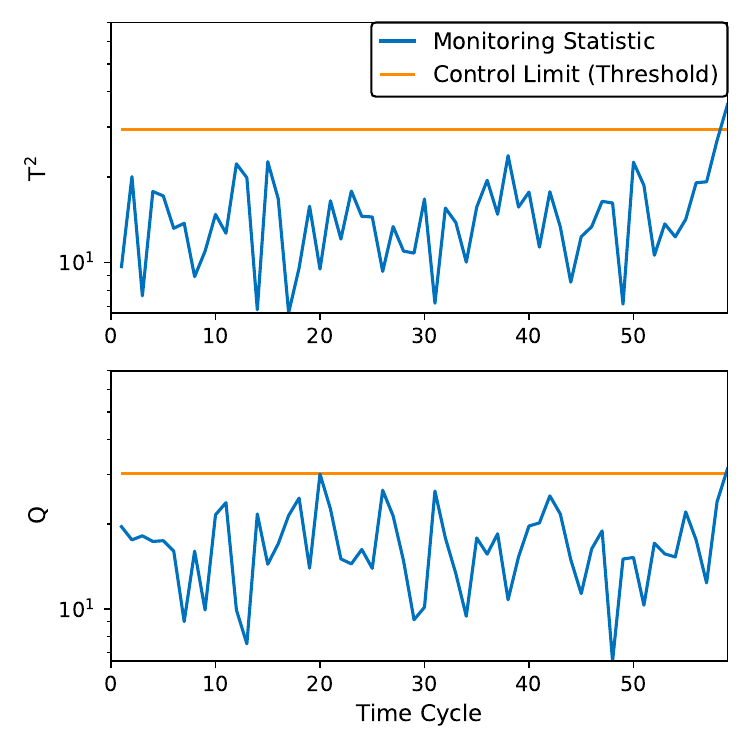}%
\label{fig_Train}}
\hfil
\subfloat[Validation cycles]{\includegraphics[width=0.5\linewidth]{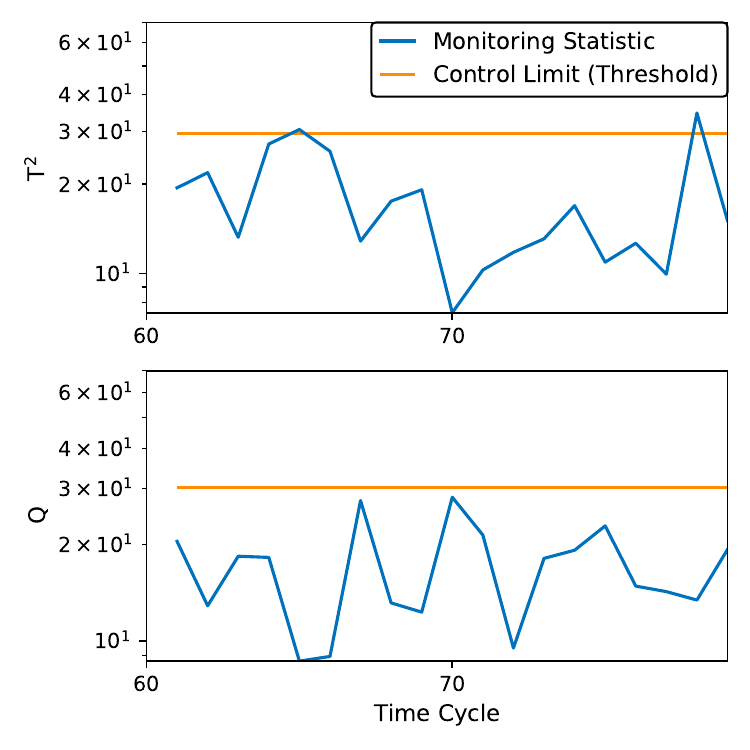}%
\label{fig_Val}}
\caption{Monitoring statistics, $T^2$ and $Q$ during normal operation with six operating conditions (a) first 60 operational cycles, and (b) next 20 operational cycles for Engine 116 of FD004.}
\label{T2Q_normaloperation}
\end{figure}

The CL of the $T^2$ and $Q$ statistics is calculated based on a 99\% confidence interval. When the monitoring statistics are consistently above the CL for normal operation, we deduce that degradation has begun. To ensure that the first 60 cycles indeed fall under normal operation, the next 20 cycles are taken as validation data and monitored against the previously calculated CL threshold. Using Engine 116 as an example from the most complex FD004 dataset, Figs. \ref{T2Q_normaloperation}\subref{fig_Train} and \ref{T2Q_normaloperation}\subref{fig_Val} show that the $T^2$ and $Q$ statistics for training and validation operational cycles are mostly below the CL with a 99\% confidence bound, which is expected from operating data in normal operation. This pattern of the $T^2$ and $Q$ statistics falling within the 99\% CL during validation cycles was observed for all train engines studied as well, leading to the reasonable conclusion that the first 80 cycles represent a state of normal operation.

\begin{figure*}[!t]
\centering
\subfloat[Engine 118 from FD002]{\includegraphics[width=0.5\linewidth]{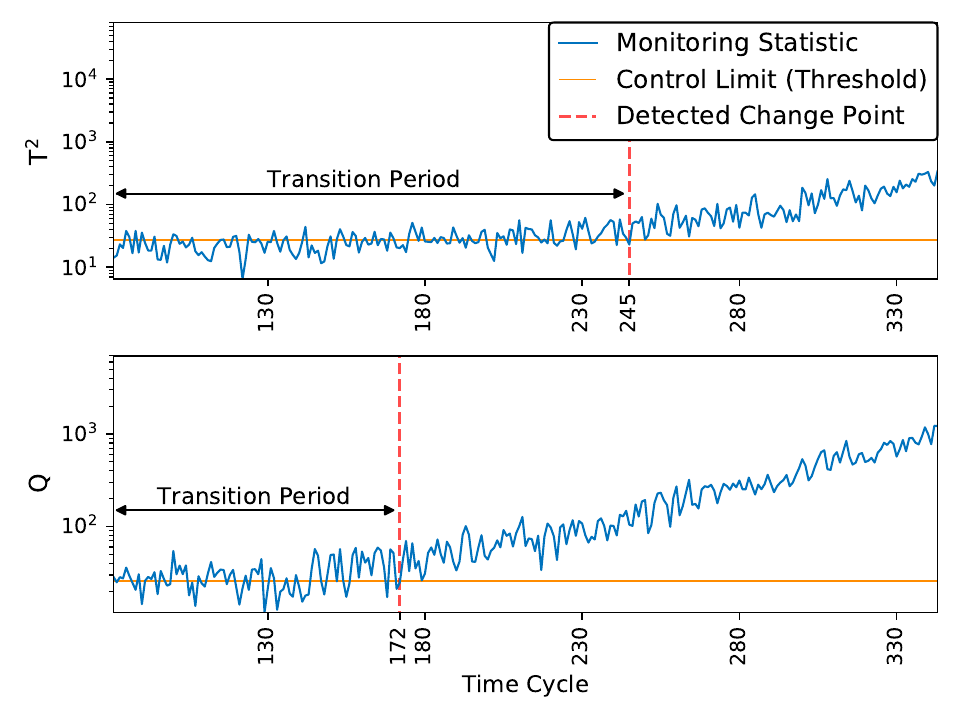}%
\label{fd02}}
\hfil
\quad
\subfloat[Engine 86 from FD003]{\includegraphics[width=0.5\linewidth]{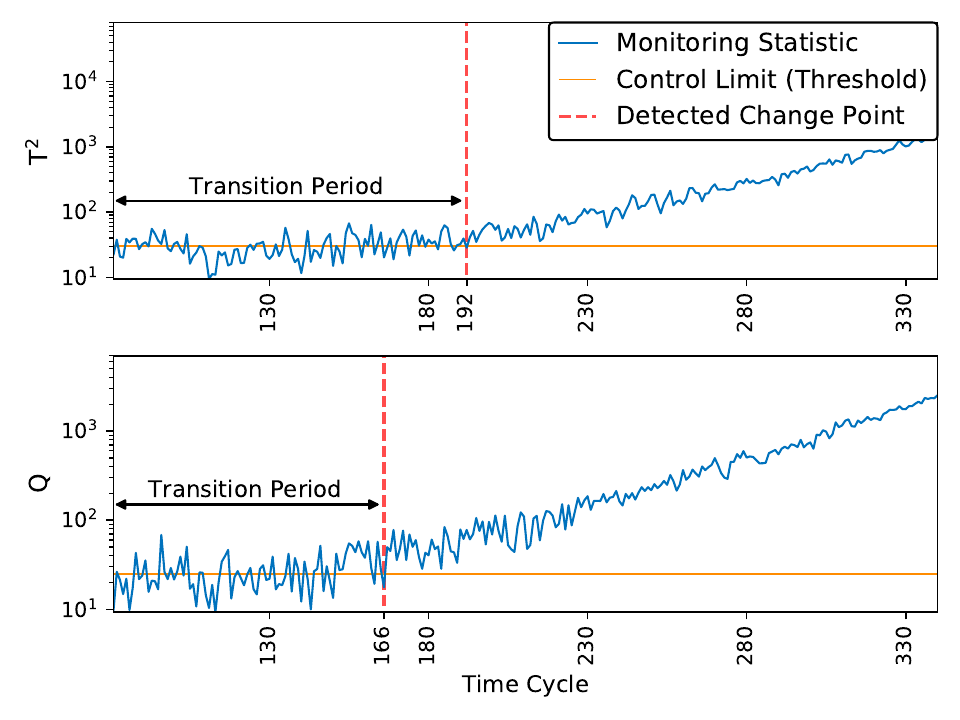}%
\label{fd03}}
\hfil
\quad
\subfloat[Engine 116 from FD004]{\includegraphics[width=0.5\linewidth]{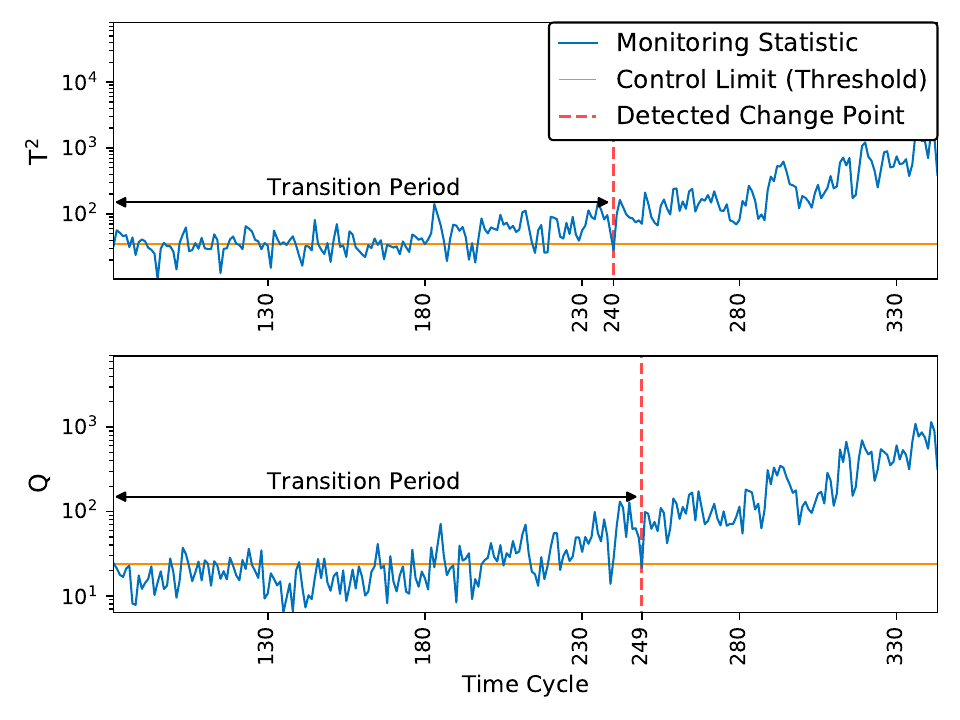}%
\label{fd04}}
\caption{Monitoring statistics, $T^2$ and $Q$ during remaining operational cycles till end of life (test cycles) for (a) Engine 118 of FD002, (b) Engine 86 of FD003, and (c) Engine 116 of FD004.}
\vspace{-1.3em}
\label{T2Q_testoperation}
\end{figure*}

\begin{figure}[!t]
\centering
\includegraphics[width=0.8\linewidth]{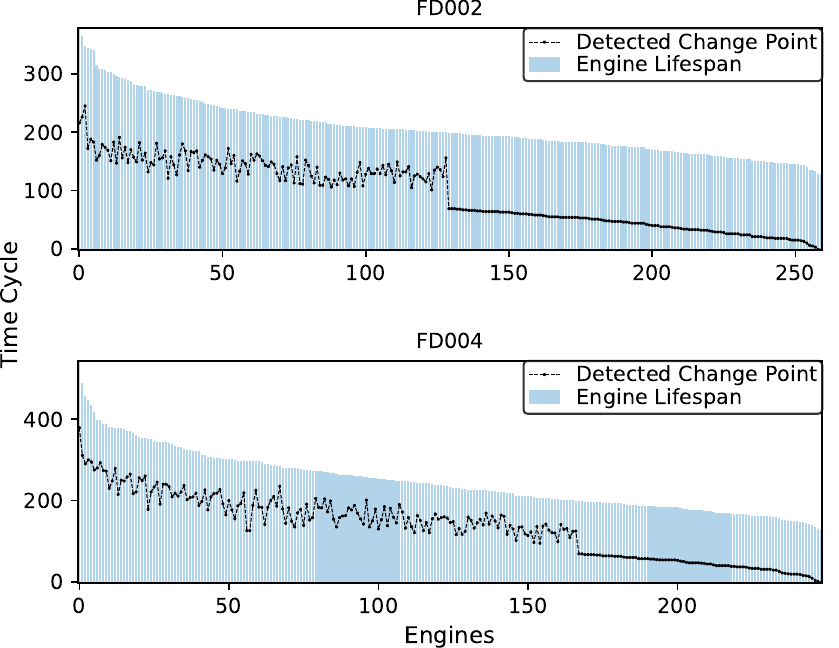}
\vspace{-1em}
\caption{Detected change point of engines against engine lifespan.}
\label{changepoint}
\end{figure} 

\subsubsection{Unsupervised Change Point Detection}
Remaining operational cycles after the first 80 cycles (60 training and 20 validation cycles) are used as testing cycles containing a yet to be determined change point at which the engine begins to degrade. Fig. \ref{T2Q_testoperation} plots the testing cycles of three engines selected from FD002 to FD004, which have multiple operating conditions and/or multiple fault modes. Engines of similar lifespan (341 to 344 cycles) were chosen for a fair comparison. Interestingly, the monitoring statistics $T^2$ and $Q$ fluctuate about the CL threshold for some time, previously termed the transition period in Section III-A2, before breaching the CL and increasing steadily away from it. 

The change point is the latest time cycle at which the $T^2$ and $Q$ statistics permanently breach the CL and stay above it. The $T^2$ and $Q$ statistics each yield a change point. The earlier change point of the two values is used to inform the subsequent labelling of the piecewise RUL function for the LSTM-based RUL estimation. Taking Engine 116 of FD004 in Fig. \ref{T2Q_testoperation}\subref{fd04} as an example, the earlier change point, 240 is selected over the later one, 249. It is reasonable to choose the earlier change point as early warning is preferred in practice to take timely preventive maintenance. It is also worth noting from Fig. \ref{T2Q_testoperation} that although the three engines selected have similar lifespan, their change points detected are appreciably different due to the varying operating conditions and fault types.

The calculated change points of each engine in the FD002 and FD004 datasets are plotted against the backdrop of their lifespans in Fig. \ref{changepoint}. As expected, the change point occurs at later cycles (indicated by larger change point values), for engines with longer lifespans. For engines with a lifespan less than 200 cycles, their change points are monotonically decreasing with respect to their lifespans, because we use a fixed upper RUL limit of 130 cycles as explained earlier.

\subsection{Change Point Integrated RUL Estimation using LSTM}
\subsubsection{Data Preparation}
Before the online RUL estimation, the change point-informed data processing steps discussed in Sections III-B1 and III-B2 are performed. First, RUL labels of the train engines are processed as a piecewise function.

The test engine dataset forms the query devices used for online RUL estimation and performance evaluation. Since the train engine RUL labels were processed in a piecewise manner, the test engine labels are similarly capped with an upper limit. As the C-MAPSS test engine dataset is a static snapshot of sensor data  up till a random time before failure, it is not possible to determine change points for the test engines using the online model in Section III-C as it would be the case for a live and continuous stream of query samples. Thus, the literature value of 130 cycles is adopted for the upper RUL limit for the true and estimated RUL of the test engines following \cite{xiang2021multicellular}. Finally, the sensor data of both train and test engines are scaled by the change point-informed piecewise standardisation described in Section III-B2.

\subsubsection{RUL Estimation}
\begin{table*}{}
\newcolumntype{P}[1]{>{\centering\arraybackslash}p{#1}}
\renewcommand{\arraystretch}{1.3}
\scriptsize
\centering
\caption{Hyperparameter selection for LSTM-based RUL estimation model.}
\label{table_hyperparameter_search}
\begin{threeparttable}
\begin{tabular}{p{2.2cm}P{2.7cm}P{1.6cm}P{1.6cm}P{1.6cm}P{1.6cm}}
\hline 
\hline
Hyperparameters & Search space & FD001 & FD002 & FD003 & FD004 \\
\hline Sequence length & $\{30,40,50\}$ & 50 & 50 & 50& 50\\
LSTM layers & $\{1,2,3\}$ & 3 & 3 &3 &3 \\
Hidden neurons & $\{32,64,100,128,256\}$ & (256,128,32) & (256,128,32) & (256,100,32) & (256,100,32)\\
Dropout ratio & $\{0,0.1,0.2\}$ & (0.2, 0.1) & (0.1, 0.1) & (0.2, 0.1) & (0.1, 0.1)\\
Learning rate & $\{0.01,0.001\}$ & $0.001$ & $0.001$ & $0.001$ & $0.001$ \\
Optimiser & RMSProp, Adam & RMSProp & RMSProp & RMSProp & RMSProp \\
\hline
\hline
\end{tabular}
\end{threeparttable}
\end{table*}

For the LSTM network for RUL estimation, there are already a large number of papers discussing the best architectures for the C-MAPSS dataset. We use the wealth of information available as a start, and verify it with our own testing to construct the our LSTM network with optimised parameters. There are several key hyperparameters that affect the RUL estimation performance. Table \ref{table_hyperparameter_search} summarises the search space considered for the hyperparameter values and the selected hyperparameter configurations for FD001 to FD004. We discuss some of the notable hyperparameters below.

First, the input sequence length (i.e., the maximum time steps fed to the LSTM cell) is an important factor affecting the learning of temporal dependencies, and consequently, the RUL estimation performance. While longer sequences can provide more contextual information to the model, there is also a risk of model overfitting if sequences are too complex relative to the available training data size \cite{kim2020state}. Furthermore, the optimal sequence length is often specific to the dataset and learning task \cite{kim2020state, zheng2017long}. Thus, we assessed a reasonable range of candidate sequence length $L \in \{30,40, 50\}$ and $L=50$ yielded the best RUL estimation for FD001 to FD004. 

The next set of hyperparameters, the number of stacked LSTM layers and the number of hidden neurons in a layer, define the LSTM network. Generally, as the number of LSTM layers (i.e., depth of the network) increases, the model's ability to learn more complex, latent relationships between feature variables increases, and thus, the RUL estimation performance may increase if it depends on these relationships. However, adding layers beyond a certain point eventually erodes model performance due to issues such as model overfitting and vanishing gradients in the backpropagation process \cite{zhang2018deep}. For our work, a 3-layer LSTM model yielded the best RUL estimation performance. We discuss the impact of the number of layers on the RUL estimation performance as a sensitivity analysis later in Section \ref{section_sensitivity_analysis}, and focus here on the role of the optimiser in combating challenges such as vanishing gradients, appropriate learning rates, and slow convergence in deep networks. We assessed both RMSProp \cite{tieleman2012rmsprop} and Adam \cite{KingBa15} as optimisers. In our experiments, RMSProp, with its ability to adaptively tune the learning rates for the model parameters based on historical gradient information, was found to yield better RUL estimation performance and faster convergence (in 30 epochs).

For the number of hidden neurons, Bengio \textit{et al.} \cite{bengio2012practical} recommends an overcomplete first hidden layer (i.e., a size larger than the input vector dimension) for better generalisability and model performance. Thus, we start off with 256 hidden neurons in the first layer, consistent with \cite{wu2018remaining} and try different configurations for the remaining layers as shown in Table \ref{table_hyperparameter_search}. Nonetheless, as the number of model parameters increases with deep layers, model overfitting becomes a concern. Thus, we add dropout \cite{baldi2013understanding} layers in between the LSTM layers as an important regularisation technique to prevent model overfitting and enhance its generalisability. The dropout ratio specifies the portion of hidden neurons to be randomly dropped out (i.e., excluded) during the training process, thus, inducing the remaining neurons to learn the needed representations for the predictions independent of the randomly dropped neurons. As seen in Table \ref{table_hyperparameter_search}, candidate dropout ratios considered were $\{0, 0.1, 0.2\}$ , where 0 denotes no dropout. The models for all datasets from FD001 to FD004 benefited from the inclusion of dropout, corroborating the crucial role of regularisation in deep networks. For example, the LSTM network for FD001 performed best with a dropout ratio of 0.2 for the first layer and 0.1 for the second layer.

To complete the RUL estimation model, the LSTM layers are combined with a fully connected output layer that decodes the learnt feature representations into a predicted RUL value. The model was trained for 30 epochs as it was sufficient to achieve good convergence and prediction performance on test engines. The performance of the RUL estimation model is evaluated based on two commonly used benchmark metrics, Root Mean Square Error (\textit{RMSE}) and, the Score Function (\textit{SF}) \cite{saxena2008damage}. The \textit{SF} is asymmetric and gives a larger penalty for overestimating the RUL as this can lead to delayed maintenance or even system failure.

\begin{table*}
\newcolumntype{P}[1]{>{\centering\arraybackslash}p{#1}}
\renewcommand{\arraystretch}{1.3}
\tiny
\centering
\caption{Performance comparison between proposed method and existing algorithms (best in \textbf{bold}, second-best \underline{underlined}).}
\vspace{2mm}
\label{Table_performance_comp}
\begin{threeparttable}
\begin{tabular}{P{1.5cm}P{2.35cm}P{0.01cm}P{0.6cm}P{0.95cm}P{0.01cm}P{0.6cm}P{0.9cm}P{0.01cm}P{0.6cm}P{0.9cm}P{0.01cm}P{0.6cm}P{0.9cm}}
\hline
\hline
\multirow{2}{1.5cm}{\centering Type}                   & \multirow{2}{2.3cm}{\centering Method} && \multicolumn{2}{c}{FD001} && \multicolumn{2}{c}{FD002} && \multicolumn{2}{c}{FD003} && \multicolumn{2}{c}{FD004} \\
\cline{4-5} \cline{7-8}\cline{10-11}\cline{13-14}
 &  && \centering\textit{RMSE}        & \centering\textit{SF}  &      & \textit{RMSE}       & \textit{SF}        & &\textit{RMSE}       & \textit{SF} & &\textit{RMSE}        & \textit{SF}      \\
\hline
\multirow{3}{1.5cm}{\centering Conventional Regressors} & RF \cite{zhang2016multiobjective}&& 17.91        & 479.75         & & 29.59        & 70456.86         & & 20.27        & 711.13         & & 31.12        & 46567.63          \\
                                           & LASSO\cite{zhang2016multiobjective} && 19.74        & 653.85       & & 37.13        & 276923.89         & & 21.38       & 1058.36         & & 40.70        & 125297.19          \\
                                           & XGBoost\cite{li2018light,ma2020interpretability} && 15.26        & 343.60         & & NA\tnote{2}        & NA\tnote{2} & & 19.33        & 943.76 & & NA\tnote{2}        & NA\tnote{2}          \\
                                           \hline
\multirow{11}{1.5cm}{\centering LSTM-based Deep Learning}                    & LSTM \cite{zheng2017long} && 16.14 & 338.00         & & 24.49        & 4450.00         & & 16.18        & 852.00         & & 28.17        & 5500.00          \\
 & A-LSTM \cite{chen2020machine} && 14.53       & 322.44         & & NA\tnote{2}         & NA\tnote{2}  & & NA\tnote{2}         & NA\tnote{2}          & & 27.08 & 5649.14          \\
                                           & Bi-LSTM \cite{huang2019bidirectional}  && NA\tnote{2}         & NA\tnote{2}          & & 25.11        & 4793.00         & & NA\tnote{2}  & NA\tnote{2}          & & 26.61        & 4971.00          \\
                                           & BS-LSTM \cite{liao2018uncertainty} && 14.89 & 481.10         & & 26.86        & 7982.00         & & 15.11        & 493.40         & & 27.11       & 5200.00          \\
                                           & CNN-LSTM \cite{wu2019weighted} && 14.40        & 290.00         & & 27.23        & 9869.00         & & 14.32       & 316.00         & & 26.69       &6594.00          \\
                                           & MC-LSTM\cite{xiang2021multicellular} && 13.71 & 315.00      & & NA\tnote{2} & NA\tnote{2} & & NA\tnote{2}        & NA\tnote{2} & & 23.81       & 4826.00        \\
                                            & Cap-LSTM\cite{zhao2021novel} && \textbf{12.27} & \underline{260.00} & & 17.79 & \underline{1850.00} & & \underline{12.55}       & \underline{217.00}  & & 22.05       & 4570.00         \\
                                           & Att-LSTM \cite{boujamza2022attention} && 13.95       & 320.00 & & \underline{17.65}        & 2102.00 & & 12.72     & 223.00 & & \underline{20.21}        & 3100.00 \\
& GA-CNN-LSTM \cite{kumar2021remaining} && 15.92       & NA\tnote{2}      & & 22.87        & NA\tnote{2}         & & 17.26 & NA\tnote{2}         & & 26.32        & NA\tnote{2}          \\
                                           & GM-LSTM \cite{sayah2021deep}                                         && 14.08        & 308.00 & & 18.59        & 1880.00 & & \textbf{12.15}       & 221.00 & & 20.91       & \underline{2633.00} \\
                                           & \textit{ChangePoint-LSTM (Ours)} && \multirow{1}{*}{\underline{13.59}}     & \multirow{1}{*}{\textbf{224.88}}        & & \multirow{1}{*}{\textbf{16.67}} & \multirow{1}{*}{\textbf{947.99}}        & & \multirow{1}{*}{12.94} & \multirow{1}{*}{\textbf{207.10}}         & & \multirow{1}{*}{\textbf{18.69}}       & \multirow{1}{*}{\textbf{1360.34}} \\
\cline{2-14}
& Improvement\tnote{1} && -      & 13.51\%       & & 5.55\% & 48.76\%         & & - & 4.56\%         & & 7.52\%      & 48.33\% \\
                          
\hline                                       \hline
\end{tabular}
\begin{tablenotes}
\item[1] The improvement calculated by comparing proposed model performance against the best-performing benchmark.
\item[2] NA is short for not applicable as the results are not provided in cited paper.
\end{tablenotes}
\end{threeparttable}
\end{table*}

\subsubsection{RUL Estimation Performance}
In this section, we evaluate the RUL estimation performance of our change point integrated model against an extensive set of benchmarks. These benchmarks range from vanilla LSTM\cite{zheng2017long} to its state-of-the-art variations\cite{xiang2021multicellular, zhao2021novel}, but, all still employ fixed literature values for the piecewise RUL construction. The results, in terms of \textit{RMSE} and \textit{SF} metrics, and the percentage improvement over the best-performing benchmarks are presented in Table \ref{Table_performance_comp}. Given the standardisation of the comparison to solely LSTM-based deep learning models, the analysis can also be likened to an ablation study, with results highlighting the impact of accounting for heterogeneous change points in RUL estimation.

As seen from Table \ref{Table_performance_comp}, our model’s performance is extremely competitive, even against recent state-of-the-art variations of LSTM. Although our model only utilises a vanilla LSTM architecture, it consistently outperforms other advanced LSTM benchmarks, in terms of \textit{SF}, a metric of greater practical significance due to its higher penalty for overestimating the RUL. Furthermore, our model's notable outperformance, in terms of both \textit{RMSE} and \textit{SF}, for the more complex FD002 and FD004 datasets suggests that factoring in individual change points before RUL estimation is especially
important for devices working under variable operating conditions. For these devices, utilising fixed literature values for the upper RUL limit may be inadequate
in capturing the complex degradation processes occurring under variable operating conditions. Instead, our results suggest that accounting for heterogeneous
starting points of degradation is crucial for achieving accurate and reliable RUL estimation.
 
\begin{figure}[!t]
\centering
\includegraphics[width=0.8\linewidth]{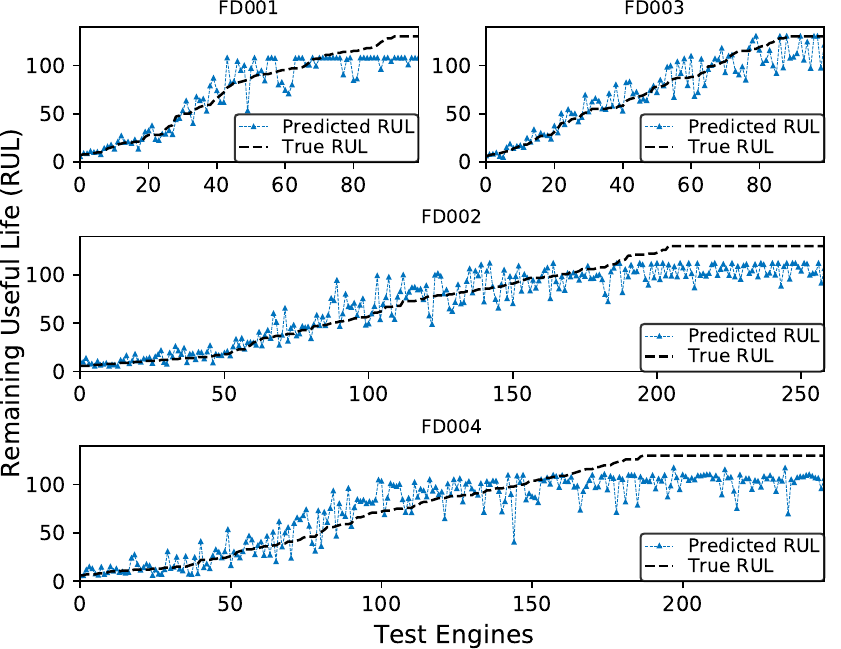}
\vspace{-0.9em}
\caption{Comparison between predicted RUL and the ground truth RUL.}
\label{RulpredVSRulTrue}
\end{figure}

Fig. \ref{RulpredVSRulTrue} plots the predicted RUL against the true RUL of each engine for an in-depth look into the prediction performance. Particularly, for FD003, the RUL prediction does well for both engines in early operation cycles (true RUL values are large) and engines in later operation cycles (true RUL values are small). For the more complex FD002 and FD004, the RUL prediction is better for engines in later operation cycles. For engines in earlier operation cycles, the predicted RUL tends to be conservatively less than the actual RUL. Nonetheless, the results are very promising given the engines are operating in a complex situation of multiple operating conditions.

\subsubsection{Sensitivity Analysis}
\label{section_sensitivity_analysis}
The RUL estimation performance is influenced by the quality of the detected change points and the  LSTM network's capacity to learn complex feature relationships. In this section, we examine the impact of two key hyperparameters on the RUL estimation performance: the number of system canonical variates $r$ (which indirectly determines the change point) and the number of LSTM layers (an indicator of the depth and learning capacity of the LSTM network).

As described in Section III-A1, the choice of $r$ directly determines the magnitude of the monitoring statistic $T^2$ and the resultant control limit $CL_{T^2}$ to detect change points, and indirectly dictates the magnitude of the $Q$ statistic and its control limit $CL_{Q}$ through the remaining residual variates. In existing literature, the optimal $r$ is typically determined based on performance of the downstream learning task \cite{odiowei2009nonlinear, ruiz2015statistical}. For example, Ruiz-C{\'a}rcel \textit{et al.} \cite{ruiz2015statistical} select $r$ for their fault detection task based on the false alarm rate. For our work, as we leverage CVA in a non-traditional manner to detect device-level change points and enhance RUL estimation, we select $r$ based on the RUL estimation performance.

We assess the \textit{RMSE} and \textit{SF} for the RUL estimation produced from the change point integrated model for a reasonable range of candidate $r \in [10,25]$, as shown in Fig. 9. The value of $r$ yielding the best RUL estimation is selected as the optimal $r$. Generally, we observe that there is no clear-cut relationship between the $r$ need for good RUL estimation performance and the presence of data complexities such as multiple fault modes or operating conditions. For instance, for both FD001 and FD003 (which differ only in terms of the number of fault modes present), the optimal value of $r$ for achieving the best RUL estimation performance was found to be 15. Similarly, the optimal $r$ for FD002 (which has a single fault mode but multiple operating conditions) was also 15. However, for the most complex dataset FD004, containing both interactions from multiple fault modes and operating conditions, the optimal $r$ achieving the best RUL estimation was larger at $r=21$. A likely reason for this lack of clear-cut relationship is because  temporal variations caused by different fault modes or operating conditions can manifest in either system space captured by $r$ or the ``noisier" residual space, depending on characteristic of the fault, the operating condition, and the potential interactions between them. However, a key advantageous aspect of our model is that we do not need in-depth domain knowledge of the fault characteristics or the operating conditions to account for them in the change point detection. As we monitor for breaches in control limits of both the $T^2$ and $Q$	statistics and conservatively utilise the earlier change point of the two, we can account for  significant changes in temporal variations in both the system space and the residual space, regardless of the fault type or operating condition.

\begin{figure}[!t]
\centering
\includegraphics[width=0.8\linewidth]{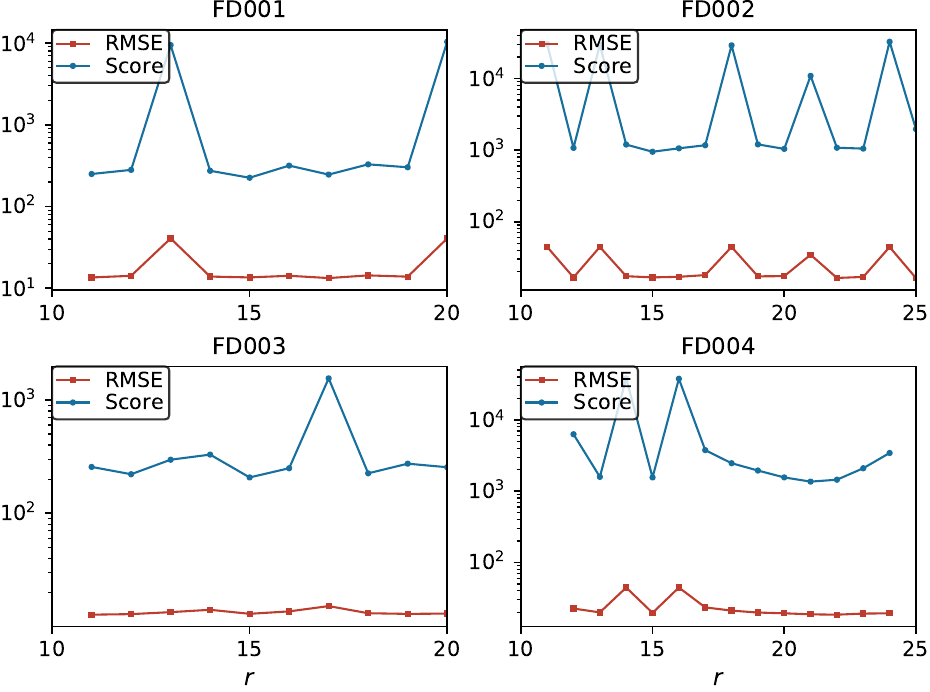}
\vspace{-0.9em}
\caption{Impact of number of system canonical variates, $r$ on \textit{RMSE} and \textit{SF}.}
\label{fig_sensitivity_r}
\end{figure}

\begin{figure}[!t]
\centering
\includegraphics[width=0.4\linewidth]{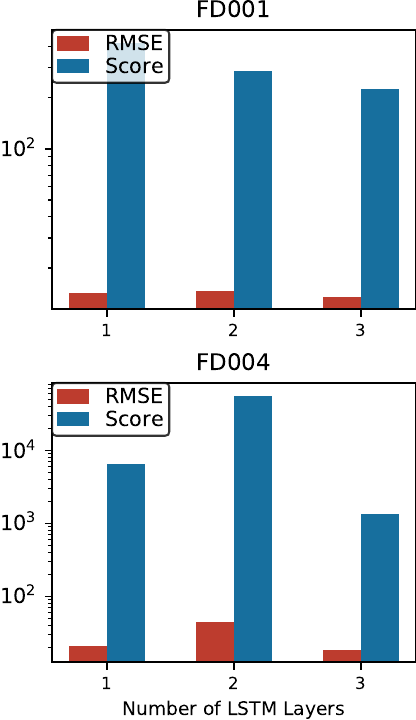}
\vspace{-0.9em}
\caption{Impact of number of LSTM layers on \textit{RMSE} and \textit{SF}.}
\label{fig_sensitivity_lstm layers}
\end{figure}

Next, we examine the impact of the number of LSTM layers on the RUL estimation performance. Generally, increasing the number of LSTM layers (i.e., depth of the network) increases the model's capacity to learn complex hierarchical relationships and dependencies of the data. However, excessive model complexity should be avoided to mitigate the risk of model overfitting. Generally, 2 to 3 LSTM layers are recommended for the C-MAPSS dataset \cite{zheng2017long, shi2021dual,zhang2018deep}. We assessed the RUL estimation performance for a candidate number of layers in $\{1, 2, 3\}$, and found that the 3-layer configuration yielded the lowest \textit{RMSE} and \textit{SF}, as shown in Fig. 10. For the sake of brevity, only the simplest dataset, FD001 and the most complex dataset, FD004 are plotted, but the conclusion on the 3-layer configuration applies to all four datasets.

\subsection{Complexity Analysis}
The two key components of our model are CVA-based change point detection and LSTM-based RUL estimation. As seen from Eq. (4), the workhorse of CVA is singular value decomposition (SVD). For the SVD computation, the flop counts (an indication of algorithm speed) increases by $\mathcal{O}\left(N \cdot p+p^3\right)$ \cite{chiang2000fault, larimore1997canonical}, where $N$ and $p$ are, respectively, the number of observations and number of time lags as defined earlier. The storage space complexity grows by $\mathcal{O}\left(N+p^2\right)$ \cite{chiang2000fault, larimore1997canonical}. Next, we discuss the computational complexity of the vanilla LSTM network used in our model. As the LSTM algorithm is local in time and space (i.e., the output of an LSTM cell depends only on the previous cell’s output and the current input), the complexity per weight and time step for updating
model weights during training is $\mathcal{O}\left(1\right)$\cite{hochreiter1997long}. Given $w$ number of weights, the complexity is thus $\mathcal{O}\left(w\right)$.

In comparison, the benchmark models that employ fixed literature values for the piecewise RUL construction can save on the computational complexity associated with a CVA-based change point detection. However, they need to compensate for this “one-size-fits-all” approach with advanced variants of LSTM and hybrid architectures to reach competitive RUL estimation results. These variants also introduce an additional layer of computational complexity to the vanilla LSTM network. For example, the  additional complexity of a convolutional layer in GA-CNN-LSTM \cite{kumar2021remaining} is $\mathcal{O}\left(s\cdot L\cdot d^2\right)$ \cite{vaswani2017attention}, where $s$, $L$, and $d$ are, respectively, the kernel size, sequence length, and the feature representation dimension. Meanwhile, in MC-LSTM \cite{xiang2021multicellular}, the additional complexity of the attention layer \cite{vaswani2017attention}  is $\mathcal{O}\left( L^2\cdot d\right)$ \cite{vaswani2017attention}.

Overall, it is evident that all models considered have to incur additional computational complexity beyond vanilla LSTM networks to learn effectively from complex datasets. However, in order to realize the significant practical benefits of change point integrated RUL estimation, we are mindful about restricting the complexity of CVA. For instance, as detailed in Section III-A1, we focus on analysing local temporal dynamics between a limited number of past and future time lags. Thus, $p \ll N$ in the complexity formulation. There are also a growing number of algorithms (e.g.,  \cite{fu2016efficient}) aimed at reducing the complexity of CVA, which could serve as a basis for the future iterations of our proposed model. 

\section{Conclusion}
\label{conc}
This paper argues that health status evaluation and change point detection are critical steps for boosting existing RUL estimation model capabilities. In our temporal learning model, we introduce a novel leveraging of canonical variate analysis for degradation monitoring and detecting device-level change points even under varying operating conditions. The proposed method of combining change point detection with LSTM-based RUL estimation outperforms existing models that do not consider heterogeneous change points, especially for Score Function values. Although turbofan engines are used as a case study, the proposed method can be easily generalised to other applications, and be combined with other deep learning RUL estimation models as it is data-driven and does not rely on domain knowledge. Future research will be directed towards extending our unsupervised change point detection methodology to account for shorter lifespan devices with less training data, and further investigating the transition period observed before the degradation state to refine the change points detected.

\bibliographystyle{IEEEtran}
\bibliography{Changepoint_preprint.bib}

\appendices
\section{Comparison of current work against existing literature on change point-informed RUL estimation}
\label{appendix_section_shi_chehade}

See table \ref{table_comparison with shi_chehade}.

\begin{table*} [h]
\newcolumntype{P}[1]{>{\centering\arraybackslash}p{#1}}
\renewcommand{\arraystretch}{1.3}
\scriptsize
\centering
\caption{Comparison of our work with existing literature on change point-informed RUL estimation.}
\label{table_comparison with shi_chehade}
\begin{threeparttable}
\begin{tabular}{|m{2.5cm}|m{2.8cm}|m{2.1cm}|m{1.4cm}|m{5cm}|}
\hline 
 & Shi and Chehade \cite{shi2021dual} & \centering Wu \textit{et al.} \cite{wu2018remaining}   & \centering Ours  & Remarks                                                                                                                                                                                                                                                                                        
 \\
 \hline
Are the standard C-MAPSS test datasets used for evaluation?    &  \centering x                      & \centering x & \centering \checkmark    & Existing works do not use the standard predefined test datasets, which are more challenging as the sensor data is available up to only some abrupt time before failure and the lifespan information is unknown. Instead, existing works use a portion of the train engines (with the full lifespan information known) for testing. \\
\hline
Is the RUL estimation performed independent of  the equipment's lifecycle stage? &  \centering x                        &  \centering x  & \centering \checkmark     & Existing works limit their RUL estimation and evaluation to only the last 50 cycles before failure. However, we perform RUL estimation on all test engines of various lifecycle stages and do not restrict our evaluation to only late-stage RUL estimation.                                                                                                             \\
\hline
Is the RUL estimation evaluated under multiple operating conditions?   & \centering x                       &  \centering \checkmark  &  \centering \checkmark     & Shi and Chehade \cite{shi2021dual} evaluate their method on the FD001 and FD003 datasets, which only have a single operating condition.                                                                                                                                                                    \\
\hline
Size of test data used for evaluation                                  & \centering Small (10 engines)      & \centering Small (20 engines) & \centering Large & Our work utilises the full, standard, pre-defined test engine dataset.                                                                                                                                                                                                                                              \\
\hline
Are standard performance evaluation metrics used?               &  \centering \checkmark                        & \centering x                  &  \centering \checkmark     & Instead of the RMSE metric, Wu \textit{et al.} \cite{wu2018remaining} report the relative prediction error, which refer to percentages of samples in the testing set with   relative prediction errors less than or equal to 5\%, 10\%, and 20\%   respectively.                                                                 \\
\hline
Is the analysis of change points detected interpretable?                 & \centering x                       & \centering x                  &  \centering \checkmark      & Our change point detection method is   highly interpretable as the monitoring statistics and departures from control   limits can be easily visualised and monitored.                                                                                                                          \\
\hline
\end{tabular}
\end{threeparttable}
\end{table*}

\section{Investigation on minimum lifespan of train engines}
\label{appendix_section_lifespan}

As discussed in Section IV-A1, the choice of the minimum lifespan needed for train engines of the change point detection model influences the CLs learnt, change points detected, and ultimately the RUL estimation. We consider a range of candidate minimum lifespans $\{100,125,150,175,200, 225\}$ to assess the appropriate minimum lifespan needed for train engines to produce well-performing RUL estimations (which is a strong indicator for the quality of change points learnt).  

For the sake of brevity, the analysis focuses on the ``worst-case'' scenarios of FD001 and FD003, which have only 100 train engines to begin with. For FD001, the lifespans of the train engines range from 128 to 362 operation cycles, with an average lifespan of 206 cycles. For FD003, the lifespans range from 145 to 525 cycles, with an average lifespan of 247 cycles. Table \ref{Table_min_lifespan} presents the \textit{RMSE} and \textit{SF} values of the RUL estimations produced from the various candidate values for the minimum lifespan needed for the train engines. We observe that the RUL estimation performance generally improves as the minimum lifespan requirements for the train engines increases. At the minimum lifespan threshold of 200 cycles, there is a significant improvement in the RUL estimation performance. This confirms that the validity of our initial assumption on the first 60 operational cycles being from normal operation is indeed stronger for train engines with at least 200 operational cycles. Interestingly, increasing the minimum lifespan requirement beyond 200 cycles worsens the RUL estimation as it considerably reduces the number of suitable train engines available for training the change point detection model.

\begin{table*}[h]
\centering
\newcolumntype{P}[1]{>{\centering\arraybackslash}p{#1}}
\renewcommand{\arraystretch}{1.3}
\scriptsize
\centering
\caption{Impact of minimum engine lifespan chosen on downstream RUL estimation (best in \textbf{bold}).}
\vspace{2mm}
\label{Table_min_lifespan}
\begin{threeparttable}
\begin{tabular}{P{2cm}P{0.01cm}P{0.7cm}P{1cm}P{0.01cm}P{0.7cm}P{1cm}}
\hline
\hline
\multirow{2}{2cm}{\centering Min. train engine lifespan}  &  & \multicolumn{2}{c}{FD001} &  & \multicolumn{2}{c}{FD003} \\
\cline{3-4} \cline{6-7}
 &  & \textit{RMSE} & \textit{SF}         &  & \textit{RMSE}          & \textit{SF} \\
\hline
100           &  & NA\tnote{1}            & NA\tnote{1}                     &  & 17.77            & 1197.28          \\
125           &  & NA\tnote{1}                      & NA\tnote{1}                     &  & 17.77            & 1197.28  \\
150           &  & NA\tnote{1}                                & NA\tnote{1}                               &  & 16.18            & 408.92         \\
175           &  & 15.89            & 304.10          &  & 15.43 & 502.75 \\
200 &  & \textbf{13.59}            & \textbf{224.88} &  & \textbf{12.94}                       & \textbf{207.10}   \\
225  &  & 40.69           & 11749.48 &  & 40.18 & 13031.71\\
\hline
\hline
\end{tabular}
\begin{tablenotes}
\item[1] NA indicates that the chosen minimum lifespan was too short to detect change points.
\end{tablenotes}
\end{threeparttable}
\end{table*}

\vspace{12pt}
\end{document}